\documentclass[journal,comsoc]{IEEEtran}
\usepackage{amsmath}
\usepackage{newtxmath}
\usepackage[T1]{fontenc}
\usepackage[latin9]{inputenc}
\usepackage{array}
\usepackage{url}
\usepackage{multirow}
\usepackage{graphicx}
\usepackage[unicode=true,
 bookmarks=false,
 breaklinks=false,pdfborder={0 0 1},backref=section,colorlinks=false]
 {hyperref}

\makeatletter

\newcommand{\lyxmathsym}[1]{\ifmmode\begingroup\def\b@ld{bold}
  \text{\ifx\math@version\b@ld\bfseries\fi#1}\endgroup\else#1\fi}

\providecommand{\tabularnewline}{\\}

\ifCLASSINFOpdf
\else
\fi
\makeatother

\begin{document}

\title{Recent Advances in Zero-shot Recognition}

\author{Yanwei~Fu, ~Tao~Xiang, ~Yu-Gang~Jiang, Xiangyang~Xue,~Leonid~Sigal,
and~Shaogang~Gong \thanks{Yanwei Fu and Xiangyang Xue are with the School of Data Science, Fudan
University, Shanghai, 200433, China. E-mail: \{yanweifu,xyxue\}@fudan.edu.cn;

Yu-Gang Jiang is with the School of Computer Science, Shanghai Key
Lab of Intelligent Information Processing, Fudan University. Email:
ygj@fudan.edu.cn; Yu-Gang Jiang is the corresponding author.

Leonid Sigal is with the Department of Computer Science, University
of British Columbia, BC, Canada. Email: lsigal@cs.ubc.ca;

Tao Xiang and Shaogang Gong are with the School of Electronic Engineering
and Computer Science, Queen Mary University of London, E1 4NS, UK.
Email: \{t.xiang, s.gong\}@qmul.ac.uk. } }
\maketitle
\begin{abstract}
With the recent renaissance of deep convolution neural networks, encouraging
breakthroughs have been achieved on the supervised recognition tasks,
where each class has sufficient training data and fully annotated
training data. However, to scale the recognition to a large number
of classes with few or now training samples for each class remains
an unsolved problem. One approach to scaling up the recognition is
to develop models capable of recognizing unseen categories without
any training instances, or zero-shot recognition/ learning. This article
provides a comprehensive review of existing zero-shot recognition
techniques covering various aspects ranging from representations of
models, and from datasets and evaluation settings. We also overview
related recognition tasks including one-shot and open set recognition
which can be used as natural extensions of zero-shot recognition when
limited number of class samples become available or when zero-shot
recognition is implemented in a real-world setting. Importantly, we
highlight the limitations of existing approaches and point out future
research directions in this existing new research area. 
\end{abstract}


\section{Introduction}

Humans can distinguish at least 30,000 basic object categories \cite{object_cat_1987}
and many more subordinate ones (e.g., breeds of dogs). They can also
create new categories dynamically from few examples or purely based
on high-level description. In contrast, most existing computer vision
techniques require hundreds, if not thousands, of labelled samples
for each object class in order to learn a recognition model. Inspired
by humans' ability to recognize without seeing examples, the research
area of \emph{learning to learn} or \emph{lifelong learning} \cite{chen_iccv13,lifelonglearning,Tom1995lifelong}
has received increasing interests.

These studies aim to intelligently apply previously learned knowledge
to help future recognition tasks. In particular, a major topic in
this research area is building recognition models capable of recognizing
novel visual categories that have no associated labelled training
samples (\emph{i.e.}, zero-shot learning), few training examples (\emph{i.e.}
one-shot learning), and recognizing the visual categories under an
`open-set' setting where the testing instance could belong to either
seen or unseen/novel categories.

These problems can be solved under the setting of transfer learning.
Typically, transfer learning emphasizes the transfer of knowledge
across domains, tasks, and distributions that are similar but not
the same. Transfer learning \cite{pan2009transfer_survey} refers
to the problem of applying the knowledge learned in one or more auxiliary
tasks/domains/sources to develop an effective model for a target task/domain.

To recognize zero-shot categories in the target domain, one has to
utilize the information learned from source domain. Unfortunately,
it may be difficult for existing methods of domain adaptation \cite{visual_domain_adapt}
to be directly applied on these tasks, since there are only few training
instances available on target domain. Thus the key challenge is to
learn domain-invariant and generalizable feature representation and/or
recognition models usable in the target domain.

The rest of this paper is organized as follows: We give an overview
of zero-shot recognition in Sec. \ref{sec:Overview-of-Zero-shot}.
The semantic representations and common models of zero-shot recognition
have been reviewed in Sec. \ref{sec:Semantic-Representations-in}
and Sec. \ref{sec:Models-for-Zero-shot} respectively. Next, we discuss
the recognition tasks beyond zero-shot recognition in Sec. \ref{sec:Beyond-Zero-shot-Recognition}
including generalized zero-shot recognition, open-set recognition
and one-shot recognition. The commonly used datasets are discussed
in Sec. \ref{sec:Datasets-in-Zero-shot}; and we also discuss the
problems of using these datasets to conduct zero-shot recognition.
Finally, we suggest some future research directions in Sec. \ref{sec:Future-Research-Directions}
and conclude the paper in Sec. \ref{sec:Conclusion}.

\section{Overview of Zero-shot Recognition\label{sec:Overview-of-Zero-shot}}

Zero-shot recognition can be used in a variety of research areas,
such as neural decoding from fMRI images \cite{palatucci2009zero_shot},
face verification \cite{kumar2009}, object recognition \cite{lampert13AwAPAMI},
video understanding \cite{emotion_0shot,fu2012attribsocial,liu2011action_attrib,yanweiPAMIlatentattrib},
and natural language processing \cite{Blitzer_zero-shotdomain}. The
tasks of identifying classes without any observed data is called zero-shot
learning. Specifically, in the settings of zero-shot recognition,
the recognition model should leverage training data from source/auxiliary
dataset/domain to identify the unseen target/testing dataset/domain.
Thus the main challenge of zero-shot recognition is how to generalize
the recognition models to identify the novel object categories without
accessing any labelled instances of these categories.

The key idea underpinning zero-shot recognition is to \emph{explore}
and \emph{exploit} the knowledge of how an unseen class (in target
domain) is semantically related to the seen classes (in the source
domain).  We \emph{explore} the relationship of seen and unseen classes
in Sec. \ref{sec:Semantic-Representations-in}, through the use of
intermediate-level semantic representations. These semantic representation
are typically encoded in a high dimensional vector space. The common
semantic representations include semantic attributes (Sec. \ref{subsec:Semantic-Attributes})
and semantic word vectors (Sec. \ref{sec:Generalised-Semantic-Representat}),
encoding linguistic context. The semantic representation is assumed
to be shared between the auxiliary/source and target/test dataset.
Given a pre-defined semantic representation, each class name can be
represented by an attribute vector or a semantic word vector \textendash{}
a representation termed {\em class prototype}.

Because the semantic representations are universal and shared, they
can be \emph{exploited} for knowledge transfer between the source
and target datasets (Sec. \ref{sec:Models-for-Zero-shot}), in order
to enable recognition novel unseen classes. A projection function
mapping visual features to the semantic representations is typically
learned from the auxiliary data, using an embedding model (Sec. \ref{subsec:Embedding-Models}).
Each unlabelled target class is represented in the same embedding
space using a class `prototype'. Each projected target instance is
then classified, using the recognition model, by measuring similarity
of projection to the class prototypes in the embedding space (Sec.
\ref{subsec:Recognition-models-in}). Additionally, under an open
set setting where the test instances could belong to either the source
or target categories, the instances of target sets can also be taken
as outliers of the source data; therefore novelty detection \cite{RichardNIPS13}
needs to be employed first to determine whether a testing instance
is on the manifold of source categories; and if it is not, it will
be further classified into one of the target categories.

The zero-shot recognition can be considered a type of life-long learning.
For example, when reading a description `flightless birds living
almost exclusively in Antarctica', most of us know and can recognize
that it is referring to a penguin, even though most people have never
seen a penguin in their life. In cognitive science \cite{Thrun96learningto},
studies explain that humans are able to learn new concepts by extracting
intermediate semantic representation or high-level descriptions ({\em
i.e.}, flightless, bird, living in Antarctica) and transferring knowledge
from known sources (other bird classes, {\em e.g.}, swan, canary,
cockatoo and so on) to the unknown target (penguin). That is the reason
why humans are able to understand new concepts with no (zero-shot
recognition) or only few training samples (few-shot recognition).
This ability is termed ``learning to learn\textquotedblright . 

More interestingly, humans can recognize newly created categories
from few examples or merely based on high-level description, {\em
e.g.}, they are able to easily recognize the video event named ``Germany
World Cup winner celebrations 2014\textquotedblright{} which, by definition,
did not exist before July 2014. To teach machines to recognize the
numerous visual concepts dynamically created by combining multitude
of existing concepts, one would require an exponential set of training
instances for a supervised learning approach. As such, the supervised
approach would struggle with the one-off and novel concepts such as
``Germany World Cup winner celebrations 2014\textquotedblright ,
because no positive video samples would be available before July 2014
when Germany finally beat Argentina to win the Cup. Therefore, zero-shot
recognition is crucial for recognizing dynamically created novel concepts
which are composed of new combinations of existing concepts. With
zero-shot learning, it is possible to construct a classifier for ``Germany
World Cup winner celebrations 2014\textquotedblright by transferring
knowledge from related visual concepts with ample training samples,
{\em e.g.}, `` FC Bayern Munich - Champions of Europe 2013\textquotedblright{}
and `` Spain World Cup winner celebrations 2010\textquotedblright .

\section{Semantic Representations in Zero-shot Recognition\label{sec:Semantic-Representations-in}}

In this section, we review the semantic representations used for 
zero-shot recognition. These representations can be categorized 
into two categories, \emph{namely}, semantic attributes and beyond.
We briefly review relevant papers in Table \ref{tab:Paper-summary-of}.

\subsection{Semantic Attributes\label{subsec:Semantic-Attributes}}

An attribute (\emph{e.g.}, has wings) refers to the intrinsic characteristic
that is possessed by an instance or a class (\emph{e.g.}, bird)~
(Fu \emph{et al.} \cite{fu2012attribsocial}), or indicates properties
(\emph{e.g.}, spotted) or annotations (\emph{e.g.}, has a head) of
an image or an object~(Lampert \emph{et al.} \cite{lampert13AwAPAMI}).
Attributes describe a class or an instance, in contrast to the typical
classification, which names an instance. Farhadi \emph{et al}. \cite{farhadi2009attrib_describe}
learned a richer set of attributes including parts, shape, materials
and \emph{etc}. Another commonly used methodology (\emph{e.g.}, in
human action recognition (Liu \emph{et al.} \cite{liu2011action_attrib}),
and in attribute and object-based modeling (Wang \emph{et al.} \cite{wang2011clothesattrib}))
is to take the attribute labels as latent variables on the training
dataset, {\emph{e}.g.}, in the form of a structured latent SVM model
with the objective is to minimize prediction loss. The attribute description
of an instance or a category is useful as a semantically meaningful
intermediate representation bridging a gap between low level features
and high level class concepts~(Palatucci \emph{et al.} \cite{palatucci2009zero_shot}).

The attribute learning approaches have emerged as a promising paradigm
for bridging the semantic gap and addressing data sparsity through
transferring attribute knowledge in image and video understanding
tasks. A key advantage of attribute learning is to provide an intuitive
mechanism for multi-task learning~(Salakhutdinov \emph{et al.} \cite{torralba2011app_share})
and transfer learning~(Hwang \emph{et al.} \cite{hwang2011obj_attrib}).
Particularly, attribute learning enables the learning with few or
zero instances of each class via attribute sharing, \emph{i.e.}, zero-shot
and one-shot learning. Specifically, the challenge of zero-shot recognition
is to recognize unseen visual object categories without any training
exemplars of the unseen class. This requires the knowledge transfer
of semantic information from auxiliary (seen) classes, with example
images, to unseen target classes.

Later works~(Parikh \emph{et al}.\cite{parikh2011relativeattrib},
Kovashka \emph{et al.} \cite{whittlesearch} and Berg \emph{et al.}
\cite{attrbDiscovery12ECCV}) extended the unary/binary attributes
to compound attributes, which makes them extremely useful for information
retrieval (\emph{e.g.}, by allowing complex queries such as ``Asian
women with short hair, big eyes and high cheekbones'') and identification
(\emph{e.g.}, finding an actor whose name you forgot, or an image
that you have misplaced in a large collection).

In a broader sense, the attribute can be taken as one special type
of ``subjective visual property'' \cite{robust_0shot}, which indicates
the task of estimating continuous values representing visual properties
observed in an image/video. These properties are also examples of
attributes, including image/video interestingness~\cite{imginterestingnessICCV2013,yugangVideoInteresting2013},
memorability~\cite{Isola2011NIPS,Isola2011cvpr}, aesthetic~\cite{Dhar2011cvpr},
and human-face age estimation~\cite{fu2010ageSurvey,crowdcountingKE}.
Image interestingness was studied in Gygli \emph{et al.}~\cite{imginterestingnessICCV2013},
which showed that three cues contribute the most to interestingness:
aesthetics, unusualness/novelty and general preferences; the last
of which refers to the fact that people, in general, find certain
types of scenes more interesting than others, for example, outdoor-natural
vs.~indoor-manmade. Jiang \emph{et al.~}\cite{yugangVideoInteresting2013}
evaluated different features for video interestingness prediction
from crowdsourced pairwise comparisons. ACM International Conference
on Multimedia Retrieval (ICMR) 2017 published special issue (``multimodal
understanding of subjective properties''\footnote{\url{http://www.icmr2017.ro/call-for-special-sessions-s1.php}})
on the applications of multimedia analysis for subjective property
understanding, detection and retrieval. These subjective visual properties
can be used as an intermediate representation for zero-shot recognition
as well as other visual recognition tasks, {\em e.g.}, people can
be recognized by the description of how pale their skin complexion
is and/or how chubby their face looks \cite{parikh2011relativeattrib}.
In the next subsections, we will briefly review different types of
attributes.

\subsubsection{User-defined Attributes\label{subsec:User-defined-Attributes}}

User-defined attributes are defined by human experts \cite{lampert2009zeroshot_dat,lampert13AwAPAMI},
or a concept ontology \cite{fu2012attribsocial}. Different tasks
may also necessitate and contain distinctive attributes, such as facial
and clothes attributes \cite{wang2011clothesattrib,moon_attrb,rudd2016moon,wang2016walk,datta2011face_attrib,ehrlich2016facial},
attributes of biological traits (\emph{e.g.}, age and gender) \cite{survey_of_face,facial_attrb_icmr},
product attributes (\emph{e.g.}, size, color, price) \cite{multi_task_attrib}
and 3D shape attributes \cite{3D_shape_attribute}. Such attributes
transcend the specific learning tasks and are, typically, pre-learned
independently across different categories, thus allowing transference
of knowledge \cite{whittlesearch,vaquero2009attrib_surveil,wang2009attrib_class_sal}.
Essentially, these attributes can either serve as the intermediate
representations for knowledge transfer in zero-shot, one-shot and
multi-task learning \cite{multi_task_attrib}, or be directly employed
for advanced applications, such as clothes recommendation \cite{wang2011clothesattrib}.

Ferrari \emph{et al.} \cite{ferrari2007attrib_learn} studied some
elementary properties such as colour and/or geometric pattern. From
human annotations, they proposed a generative model for learning simple
color and texture attributes. The attribute can be either viewed as
unary (\emph{e.g.}, red colour, round texture), or binary (\emph{e.g.},
black/white stripes). The `unary' attributes are simple attributes,
whose characteristic properties are captured by individual image segments
(appearance for red, shape for round). In contrast, the `binary' attributes
are more complex attributes, whose basic element is a pair of segments
(\emph{e.g.}, black/white stripes).

\subsubsection{Relative Attributes}

Attributes discussed above use single value to represent the strength
of an attribute being possessed by one instance/class; they can indicate
properties (\emph{e.g.}, spotted) or annotations of images or objects.
In contrast, relative information, in the form of relative attributes,
can be used as a more informative way to express richer semantic meaning
and thus better represent visual information. The relative attributes
can be directly used for zero-shot recognition \cite{parikh2011relativeattrib}.

Relative attributes (Parikh \emph{et al.} \cite{parikh2011relativeattrib})
were first proposed in order to learn a ranking function capable of
predicting the relative semantic strength of a given attribute. The
annotators give pairwise comparisons on images and a ranking function
is then learned to estimate relative attribute values for unseen images
as ranking scores. These relative attributes are learned as a form
of richer representation, corresponding to the strength of visual
properties, and used in a number of tasks including visual recognition
with sparse data, interactive image search (Kovashka \emph{et al.}
\cite{whittlesearch}), semi-supervised (Shrivastava \emph{et al.}
\cite{ShrivastavaECCV12}) and active learning (Biswas \emph{et al.}
\cite{BiswasCVPR13,attr_clas_feedback}) of visual categories. Kovashka
\emph{et al.} \cite{whittlesearch} proposed a novel model of feedback
for image search where users can interactively adjust the properties
of exemplar images by using relative attributes in order to best match
his/her ideal queries.

Fu \emph{et al.} \cite{robust_0shot} extended the relative attributes
to ``subjective visual properties'' and proposed a learning-to-rank
model of pruning the annotation outliers/errors in crowdsourced pairwise
comparisons. Given only weakly-supervised pairwise image comparisons,
Singh \emph{et al.} \cite{relative_ranking_eccv16} developed an end-to-end
deep convolutional network to simultaneously localize and rank relative
visual attributes. The localization branch in \cite{relative_ranking_eccv16}
is adapted from the spatial transformer network \cite{jaderberg2015spatial}.

\begin{table*}
\centering{}%
\begin{tabular}{cc}
\hline 
Different Types of Attributes  & Papers\tabularnewline
\hline 
User-defined attributes  & \cite{lampert2009zeroshot_dat,lampert13AwAPAMI}\cite{fu2012attribsocial}\cite{vaquero2009attrib_surveil,wang2009attrib_class_sal}\cite{wang2011clothesattrib,moon_attrb,rudd2016moon,wang2016walk,datta2011face_attrib,ehrlich2016facial}\cite{multi_task_attrib}\cite{wang2011clothesattrib}\cite{survey_of_face,facial_attrb_icmr}\cite{ferrari2007attrib_learn}\tabularnewline
Relative attributes  & \cite{parikh2011relativeattrib}\cite{relative_ranking_eccv16}\cite{robust_0shot}\cite{BiswasCVPR13,attr_clas_feedback}\cite{whittlesearch}\cite{ShrivastavaECCV12}\tabularnewline
Data-driven attributes  & \cite{parikh2011nameable_attribs}\cite{tang2009concepts_from_noisytags}\cite{liu2011action_attrib,farhadi2009attrib_describe}\cite{fu2012attribsocial,yanweiPAMIlatentattrib}\cite{farhadi2009attrib_describe}\cite{video_story_1shot}\tabularnewline
Video attributes  & \cite{hauptmann2007semanticGapRetr}\cite{snoek2007semantic_retrieval}\cite{hauptmann2007semanticGapRetr}\cite{toderici2010youtube_tag}\cite{zuxuan_2016_CVPR,obj2action}\cite{tang2009annotation}\cite{qi2007corr_mlab}\tabularnewline
\hline 
\hline 
Concept ontology  & \cite{fergus2010label_share}\cite{RohrbachCVPR12,rohrbach2010semantic_transfer}\cite{costa_mlzsl}\cite{recog_action,concept_not_alone}\tabularnewline
Semantic word embedding  & \cite{ZSL_convex_optimization,zhang2016zero,zhang2016zeroshot,yanweiBMVC,DeviseNIPS13,RichardNIPS13}\cite{RichardNIPS13}\cite{huang2012ACL}\cite{yanweiBMVC}\cite{DeviseNIPS13,ZSL_convex_optimization}\cite{TAC_0shot,emotion_0shot}\tabularnewline
\hline 
\end{tabular}\caption{\label{tab:Paper-summary-of} Different types of semantic representations
for zero-shot recognition.}
\end{table*}

\subsubsection{Data-driven attributes}

The attributes are usually defined by extra knowledge of either expert
users or concept ontology. To better augment such user-defined attributes,
Parikh \emph{et al}. \cite{parikh2011nameable_attribs} proposed a
novel approach to actively augment the vocabulary of attributes to
both help resolve intra-class confusions of new attributes and coordinate
the ``name-ability'' and ``discriminativeness'' of candidate attributes.
However, such user-defined attributes are far from enough to model
the complex visual data. The definition process can still be either
inefficient (costing substantial effort of user experts) and/or insufficient
(descriptive properties may not be discriminative). To tackle such
problems, it is necessary to automatically discover more discriminative
intermediate representations from visual data, \emph{i.e.} data-driven
attributes. The data-driven attributes can be used in zero-shot recognition
tasks \cite{liu2011action_attrib,fu2012attribsocial}.

Despite previous efforts, an exhaustive space of attributes is unlikely
to be available, due to the expense of ontology creation, and a simple
fact that semantically obvious attributes, for humans, do not necessarily
correspond to the space of detectable and discriminative attributes.
One method of collecting labels for large scale problems is to use
Amazon Mechanical Turk (AMT) \cite{amazon_mechanical}. However, even
with excellent quality assurance, the results collected still exhibit
strong label noise. Thus label-noise \cite{tang2009concepts_from_noisytags}
is a serious issue in learning from either AMT, or existing social
meta-data. More subtly, even with an exhaustive ontology, only a subset
of concepts from the ontology are likely to have sufficient annotated
training examples, so the portion of the ontology which is effectively
usable for learning, may be much smaller. This inspired the works
of automatically mining the attributes from data.

Data-driven attributes have only been explored in a few previous works.
Liu \emph{et al}. \cite{liu2011action_attrib} employed an information
theoretic approach to infer the data-driven attributes from training
examples by building a framework based on a latent SVM formulation.
They directly extended the attribute concepts in images to comparable
``action attributes'' in order to better recognize human actions.
Attributes are used to represent human actions from videos and enable
the construction of more descriptive models for human action recognition.
They augmented user-defined attributes with data-driven attributes
to better differentiate existing classes. Farhadi \emph{et al.~}\cite{farhadi2009attrib_describe}
also learned user-defined and data-driven attributes.

The data-driven attribute works in \cite{liu2011action_attrib,farhadi2009attrib_describe,latent_semantic_attrb}
are limited. First, they learn the user-defined and data-driven attributes
separately, rather than jointly in the same framework. Therefore data-driven
attributes may re-discover the patterns that exist in the user-defined
attributes. Second, the data-driven attributes are mined from data
and we do not know the corresponding semantic attribute names for
the discovered attributes. For those reasons, usually data-driven
attributes can not be directly used in zero-shot learning. These limitations
inspired the works of \cite{fu2012attribsocial,yanweiPAMIlatentattrib}.
Fu \emph{et al.} \cite{fu2012attribsocial,yanweiPAMIlatentattrib}
addressed the tasks of understanding multimedia data with sparse and
incomplete labels. Particularly, they studied the videos of social
group activities by proposing a novel scalable probabilistic topic
model for learning a semi-latent attribute space. The learned multi-modal
semi-latent attributes can enable multi-task learning, one-shot learning
and zero-shot learning. Habibian \emph{et al.} \cite{video_story_1shot}
proposed a new type of video representation by learning the ``VideoStory''
embedding from videos and corresponding descriptions. This representation
can also be interpreted as data-driven attributes. The work won the
best paper award in ACM Multimedia 2014.

\subsubsection{Video Attributes}

Most existing studies on attributes focus on object classification
from static images. Another line of work instead investigates attributes
defined in videos, \emph{i.e.}, video attributes, which are very important
for corresponding video related tasks such as action recognition and
activity understanding. Video attributes can correspond to a wide
range of visual concepts such as objects (e.g., animal), indoor/outdoor
scenes (e.g., meeting, snow), actions (e.g. blowing candle) and events
(e.g., wedding ceremony), and so on. Compared to static image attributes,
many video attributes can only be computed from image sequences and
are more complex in that they often involve multiple objects.

Video attributes are closely related to video concept detection in
Multimedia community. The video concepts in a video ontology can be
taken as video attributes in zero-shot recognition. Depending on the
ontology and models used, many approaches on video concept detection
(Chang \emph{et al}. \cite{change_aaai,change_ijcai}, Snoek \emph{et
al.} \cite{snoek2007semantic_retrieval}, Hauptmann \emph{et al.}
\cite{hauptmann2007semanticGapRetr}, Gan \emph{et al}. \cite{recog_action}
and Qin \emph{et al.} \cite{zero_shot_action_cvpr2017}) can therefore
be seen as addressing a sub-task of video attribute learning to solve
zero-shot video event detection. Some works aim to automatically expand
(\emph{e.g.}, Hauptmann \emph{et al.} \cite{hauptmann2007semanticGapRetr}
and Tang \emph{et al.} \cite{tang2009annotation}) or enrich (Yang
\emph{et al.} \cite{yang2011tag_tagging}) the set of video tags \cite{hospedales2011video_tags,toderici2010youtube_tag,yang2011tag_tagging}
given a search query. In this case, the expanded/enriched tagging
space has to be constrained by a fixed concept ontology, which may
be very large and complex \cite{toderici2010youtube_tag,Aradhye2009,yang2011disc_subtag}.
For example, there is a vocabulary space of over $20,000$ tags in
\cite{toderici2010youtube_tag}.

Zero-shot video event detection has also attracted large research
attention recently. The video event is a higher level semantic entity
and is typically composed of multiple concepts/video attributes. For
example, a ``birthday party\textquotedblright{} event consists of
multiple concepts, \emph{e.g.}, ``blowing candle\textquotedblright{}
and ``birthday cake\textquotedblright . The semantic correlation
of video concepts has also been utilized to help predict the video
event of interest, such as weakly supervised concepts \cite{multimodal_0shot},
pairwise relationships of concepts (Gan \emph{et al.} \cite{concept_not_alone})
and general video understanding by object and scene semantics attributes
\cite{zuxuan_2016_CVPR,obj2action}. Note, a full survey of recent
works on zero-shot video event detection is beyond the scope of this
paper.

\subsection{Semantic Representations Beyond Attributes\label{sec:Generalised-Semantic-Representat}}

Besides the attributes, there are many other types of semantic representations,
\emph{e.g.} semantic word vector and concept ontology. Representations
that are directly learned from textual descriptions of categories
have also been investigated, such as Wikipedia articles \cite{Elhoseiny_2013_ICCV,deep_0shot},
sentence descriptions \cite{deep_0shot_cvpr} or knowledge graphs
\cite{RohrbachCVPR12,rohrbach2010semantic_transfer}.

\subsubsection{Concept ontology}

Concept ontology is directly used as the semantic representation alternative
to attributes. For example, WordNet~\cite{WordNet_1995Miller} is
one of the most widely studied concept ontologies. It is a large-scale
semantic ontology built from a large lexical dataset of English. Nouns,
verbs, adjectives and adverbs are grouped into sets of cognitive synonyms
(synsets) which indicate distinct concepts. The 
idea of semantic distance, defined by the WordNet ontology, is also
used by Rohrbach \emph{et al}.~\cite{RohrbachCVPR12,rohrbach2010semantic_transfer}
for transferring semantic information in zero-shot learning problems.
They thoroughly evaluated many alternatives of semantic links between
auxiliary and target classes by exploring linguistic bases such as
WordNet, Wikipedia, Yahoo Web, Yahoo Image, and Flickr Image. Additionally,
WordNet has been used for many vision problems. Fergus \emph{et al}.~\cite{fergus2010label_share}
leveraged the WordNet ontology hierarchy to define semantic distance
between any two categories for sharing labels in classification. The
COSTA \cite{costa_mlzsl} model exploits the co-occurrences of visual
concepts in images for knowledge transfer in zero-shot recognition.

\subsubsection{Semantic word vectors}

Recently, word vector approaches, based on distributed language representations,
have gained popularity in zero-shot recognition \cite{ZSL_convex_optimization,zhang2016zero,zhang2016zeroshot,yanweiBMVC,DeviseNIPS13,RichardNIPS13}.
A user-defined semantic attribute space is pre-defined and each dimension
of the space has a specific semantic meaning according to either human
experts or concept ontology ({\em e.g.}, one dimension could correspond
to `has fur', and another `has four legs')(Sec. \ref{subsec:User-defined-Attributes}).
In contrast, the semantic word vector space is trained from linguistic
knowledge bases such as Wikipedia and UMBCWebBase using natural language
processing models \cite{huang2012ACL,wordvectorICLR}. As a result,
although the relative positions of different visual concepts will
have semantic meaning, e.g., a cat would be closer to a dog than a
sofa, each dimension of the space does not have a specific semantic
meaning. The language model is used to project each class' textual
name into this space. These projections can be used as prototypes
for zero-shot learning. Socher \emph{et al}.~\cite{RichardNIPS13}
learned a neural network model to embed each image into a $50$-dimensional
word vector semantic space, which was obtained using an unsupervised
linguistic model~\cite{huang2012ACL} trained on Wikipedia text.
The images from either known or unknown classes could be mapped into
such word vectors and classified by finding the closest prototypical
linguistic word in the semantic space.

Distributed semantic word vectors have been widely used for zero-shot
recognition. Skip-gram model and CBOW model ~\cite{wordvectorICLR,distributedword2vec2013NIPS}
were trained from a large scale of text corpora to construct semantic
word space. Different from the unsupervised linguistic model~\cite{huang2012ACL},
distributed word vector representations facilitate modeling of syntactic
and semantic regularities in language and enable vector-oriented reasoning
and vector arithmetics. For example, $Vec(\lyxmathsym{\textquotedblleft}Moscow\lyxmathsym{\textquotedblright})$
should be much closer to $Vec(\lyxmathsym{\textquotedblleft}Russia\lyxmathsym{\textquotedblright})+Vec(\lyxmathsym{\textquotedblleft}capital\lyxmathsym{\textquotedblright})$
than $Vec(\lyxmathsym{\textquotedblleft}Russia\lyxmathsym{\textquotedblright})$
or $Vec(\lyxmathsym{\textquotedblleft}capital\lyxmathsym{\textquotedblright})$
in the semantic space. One possible explanation and intuition underlying
these syntactic and semantic regularities is the distributional hypothesis
\cite{Harris1981}, which states that a word's meaning is captured
by other words that co-occur with it. Frome \emph{et al}.~\cite{DeviseNIPS13}
further scaled such ideas to recognize large-scale datasets. They
proposed a deep visual-semantic embedding model to map images into
a rich semantic embedding space for large-scale zero-shot recognition.
Fu \emph{et al.}~\cite{yanweiBMVC} showed that such a reasoning
could be used to synthesize all different label combination prototypes
in the semantic space and thus is crucial for multi-label zero-shot
learning. More recent work of using semantic word embedding includes
\cite{ZSL_convex_optimization,zhang2016zero,zhang2016zeroshot}.

More interestingly, the vector arithmetics of semantic emotion word
vectors is matching the psychological theories of Emotion, such as
Ekman's six pan-cultural basic emotions or Plutchik's emotion. For
example, $Vec(\lyxmathsym{\textquotedblleft}Surprise\lyxmathsym{\textquotedblright})+Vec(\lyxmathsym{\textquotedblleft}Sadness\lyxmathsym{\textquotedblright})$
is very close to $Vec(\lyxmathsym{\textquotedblleft}Disappointment\lyxmathsym{\textquotedblright})$;
and $Vec(\lyxmathsym{\textquotedblleft}Joy\lyxmathsym{\textquotedblright})+Vec(\lyxmathsym{\textquotedblleft}Trust\lyxmathsym{\textquotedblright})$
is very close to $Vec(\lyxmathsym{\textquotedblleft}Love\lyxmathsym{\textquotedblright})$.
Since there are usually thousands of words that can describe emotions,
zero-shot emotion recognition has been also investigated in \cite{TAC_0shot}
and \cite{emotion_0shot}.

\section{Models for Zero-shot Recognition\label{sec:Models-for-Zero-shot}}

With the help of semantic representations, zero-shot recognition can
usually be solved by first learning an embedding model (Sec. \ref{subsec:Embedding-Models})
and then doing recognition (Sec. \ref{subsec:Recognition-models-in}).
To the best of our knowledge, a general `embedding' formulation of
zero-shot recognition was first introduced by Larochelle \emph{et
al.~}\cite{zero_data_AAAI2008}. They embedded handwritten character
with a typed representation which further helped to recognize unseen
classes.

The embedding models aim to establish connections between seen classes
and unseen classes by projecting the low-level features of images/videos
close to their corresponding semantic vectors (prototypes). Once the
embedding is learned, from known classes, novel classes can be recognized
based on the similarity of their prototype representations and predicted
representations of the instances in the embedding space. The recognition
model matches the projection of the image features against the unseen
class prototypes (in the embedding space). In addition to discussing
these models and recognition methods in Sec. \ref{subsec:Embedding-Models}
and Sec. \ref{subsec:Recognition-models-in}, respectively, we will
also discuss the potential problems encountered in zero-shot recognition
models in Sec. \ref{subsec:Problems-in-Existing}.

\subsection{Embedding Models \label{subsec:Embedding-Models}}

\subsubsection{Bayesian Models}

The embedding models can be learned using a Bayesian formulation,
which enables easy integration of prior knowledge of each type of
attribute to compensate for limited supervision of novel classes in
image and video understanding. A generative model is first proposed
in Ferrari and Zisserman in \cite{ferrari2007attrib_learn} for learning
simple color and texture attributes.

Lampert \emph{et al}. \cite{lampert2009zeroshot_dat,lampert13AwAPAMI}
is the first to study the problem of object recognition of categories
for which no training examples are available. Direct Attribute Prediction
(DAP) and Indirect Attribute Prediction (IAP) are the first two models
for zero-shot recognition \cite{lampert2009zeroshot_dat,lampert13AwAPAMI}.
DAP and IAP algorithms use a single model that first learns embedding
using Support Vector Machine (SVM) and then does recognition using
Bayesian formulation. The DAP and IAP further inspired later works
that employ generative models to learn the embedding, including with
topic models \cite{yanweiPAMIlatentattrib,fu2012attribsocial,yu2010attributetransfer}
and random forests \cite{Jayaraman2014}. We briefly describe the
DAP and IAP models as follows, 
\begin{itemize}
\item \emph{DAP~Model.}\quad{}Assume the relation between known classes,
$y_{i},...,y_{k}$, unseen classes, $z_{1},...,z_{L}$, and descriptive
attributes $a_{1},...,a_{M}$ is given by the matrix of binary associations
values $a_{m}^{y}$ and $a_{m}^{z}$. Such a matrix encodes the presence/absence
of each attribute in a given class. Extra knowledge is applied to
define such an association matrix, for instance, by leveraging human
experts~(Lampert \emph{et al.} \cite{lampert2009zeroshot_dat,lampert13AwAPAMI}),
by consulting a concept ontology~(Fu \emph{et al.} \cite{yanweiPAMIlatentattrib}),
or by semantic relatedness measured between class and attribute concepts~(Rohrbach
\emph{et al.} \cite{RohrbachCVPR12}). In the training stage, the
attribute classifiers are trained from the attribute annotations of
known classes $y_{i},...,y_{k}$. At the test stage, the posterior
probability $p(a_{m}|x)$ can be inferred for an individual attribute
$a_{m}$ in an image $x$. To predict the class label of object class
$z$, 
\end{itemize}
\begin{align}
p(z|x) & =\Sigma_{a\in\left\{ 0,1\right\} ^{M}}p(z|a)p(a|x)\label{eq:DAPmodel}\\
= & \frac{p(z)}{p(a^{z})}\prod_{m=1}^{M}p(a_{m}|x)^{a_{m}^{z}}
\end{align}

\begin{itemize}
\item \emph{IAP~Model.}\quad{}The DAP model directly learns attribute
classifiers from the known classes, while the IAP model builds attribute
classifiers by combining the probabilities of all associated known
classes. It is also introduced as direct similarity-based model in
Rohrbach \emph{et al.} \cite{RohrbachCVPR12}. In the training step,
we can learn the probabilistic multi-class classifier to estimate
$p(y_{k}|x)$ for all training classes $y_{i},...,y_{k}$. Once $p(a|x)$
is estimated, we use it in the same way as in for DAP in zero-shot
learning classification problems. In the testing step, we predict, 
\end{itemize}
\begin{equation}
p(a_{m}|x)=\Sigma_{k=1}^{K}p(a_{m}|y_{k})p(y_{k}|x)\label{eq:IAP model}
\end{equation}

\subsubsection{Semantic Embedding}

Semantic embedding learns the mapping from visual feature space to
the semantic space which has various semantic representations. As
discussed in Sec. \ref{subsec:Semantic-Attributes}, the attributes
are introduced to describe objects; and the learned attributes may
not be optimal for recognition tasks. To this end, Akata \emph{et
al.} \cite{labelembeddingcvpr13} proposed the idea of label embedding
that takes attribute-based image classification as a label-embedding
problem by minimising the compatibility function between an image
and a label embedding. In their work, a modified ranking objective
function was derived from the WSABIE model~\cite{WASABIE2010}. As
object-level attributes may suffer from the problems of the partial
occlusions, scale changes of images, Li \emph{et al.} \cite{LiECCV2014}
proposed learning and extracting attributes on segments containing
the entire object; and then joint learning for simultaneous object
classification and segment proposal ranking by attributes. They thus
learned the embedding by the max-margin empirical risk over both the
class label as well as the segmentation quality. Other semantic embedding
algorithms have also been investigated such as semi-supervised max-margin
learning framework \cite{max_margin_zsl_2015,sslzsl_0shot}, latent
SVM \cite{zhang2016zero} or multi-task learning \cite{hwang2011obj_attrib,decorrelated_cvpr14,unified_model}.

\subsubsection{Embedding into Common Spaces}

Besides the semantic embedding, the relationship of visual and semantic
space can be learned by jointly exploring and exploiting a common
intermediate space. Extensive efforts \cite{deep_0shot,unified_model,embedding_akata,romera2015embarrassingly,yanweiembedding,yang2014unified,mahajan2011joint_attrib}
had been made towards this direction. Akata \emph{et al.} \cite{embedding_akata}
learned a joint embedding semantic space between attributes, text
and hierarchical relationships. Ba \emph{et al.} \cite{deep_0shot}
employed text features to predict the output weights of both the convolutional
and the fully connected layers in a deep convolutional neural network
(CNN).

On one dataset, there may exist many different types of semantic representations.
Each type of representation may contain complementary information.
Fusing them can potentially improve the recognition performance. Thus
several recent works studied different methods of multi-view embedding.
Fu \emph{et al.} \cite{semantic_graph} employed the semantic class
label graph to fuse the scores of different semantic representations.
Similarly label relation graphs have also been studied in \cite{Deng2014}
and significantly improved large-scale object classification in supervised
and zero-shot recognition scenarios.

A number of successful approaches to learning a semantic embedding
space reply on Canonical Component Analysis (CCA). Hardoon \emph{et
al.}~\cite{CCAoverview} proposed a general, kernel CCA method, for
learning semantic embedding of web images and their associated text.
Such embedding enables a direct comparison between text and images.
Many more works \cite{SocherFeiFeiCVPR2010,multiviewCCAIJCV,HwangIJCV,topicimgannot}
focused on modeling the images/videos and associated text (\emph{e.g.},
tags on Flickr/YouTube). Multi-view CCA is often exploited to provide
unsupervised fusion of different modalities. Gong \emph{et al. }\cite{multiviewCCAIJCV}
also investigated the problem of modeling Internet images and associated
text or tags and proposed a three-view CCA embedding framework for
retrieval tasks. Additional view allows their framework to outperform
a number of two-view baselines on retrieval tasks. Qi \emph{et al}.
\cite{jointly_zsl} proposed an embedding model for jointly exploring
the functional relationships between text and image features for transferring
inter-model and intra-model labels to help annotate the images. The
inter-modal label transfer can be generalized to zero-shot recognition.

\subsubsection{Deep Embedding}

Most of recent zero-shot recognition models have to rely the state-of-the-art
deep convolutional models to extract the image features. As one of
the first works, DeViSE \cite{DeviseNIPS13} extended the deep architecture
to learn the visual and semantic embedding; and it can identify visual
objects using both labeled image data as well as semantic information
gleaned from unannotated text. ConSE \cite{ZSL_convex_optimization}
constructed the image embedding approach by mapping images into the
semantic embedding space via convex combination of the class label
embedding vectors. Both DeViSE and ConSE are evaluated on large-scale
datasets, \textendash{} ImageNet (ILSVRC) 2012 1K and ImageNet 2011
21K dataset.

To combine the visual and textual branches in the deep embedding,
different loss functions can be considered, including margin-based
losses \cite{DeviseNIPS13,yang2014unified}, or Euclidean distance
loss \cite{szegedy2015going}, or least square loss \cite{deep_0shot}.
Zhang \emph{et al.} \cite{deep_0shot_recent} employed the visual
space as the embedding space and proposed an end-to-end deep learning
architecture for zero-shot recognition. Their networks have two branches:
visual encoding branch which uses convolutional neural network to
encode the input image as a feature vector, and the semantic embedding
branch which encodes the input semantic representation vector of each
class which the corresponding image belonging to.

\subsection{Recognition Models in the Embedding Space\label{subsec:Recognition-models-in}}

Once the embedding model is learned, the testing instances can be
projected into this embedding space. The recognition can be carried
out by using different recognition models. The most common used one
is nearest neighbour classifier which classify the testing instances
by assigning the class label in term of the nearest distances of the
class prototypes against the projections of testing instances in the
embedding space. Fu \emph{et al.} \cite{yanweiPAMIlatentattrib} proposed
semi-latent zero-shot learning algorithm to update the class prototypes
by one step self-training.

Manifold information can be used in the recognition models in the
embedding space. Fu \emph{et al.} \cite{transductiveEmbeddingJournal}
proposed a hyper-graph structure in their multi-view embedding space;
and zero-shot recognition can be addressed by label propagation from
unseen prototype instances to unseen testing instances. Changpinyo
\emph{et al.} \cite{synthesized_0shot} synthesized classifiers in
the embedding space for zero-shot recognition. For multi-label zero-shot
learning, the recognition models have to consider the co-occurrence/correlations
of different semantic labels \cite{costa_mlzsl,yanweiBMVC,fast_0shot}.

Latent SVM structure has also been used as the recognition models
\cite{wang2010reg_tag_corr,hwang2011obj_attrib}. Wang \emph{et al.
}\cite{wang2010reg_tag_corr} treated the object attributes as latent
variables and learnt the correlations of attributes through an undirected
graphical model. Hwang \emph{et al.} \cite{hwang2011obj_attrib} utilized
a kernelized multi-task feature learning framework to learn the sharing
features between objects and their attributes. Additionally, Long
et al. \cite{shaoling_cvpr2017} employed the attributes to synthesize
unseen visual features at training stage; and thus zero-shot recognition
can be solved by the conventional supervised classification models.

\subsection{Problems in Zero-shot Recognition \label{subsec:Problems-in-Existing}}

There are two intrinsic problems in zero-shot recognition, namely
projection domain shift problem (Sec. \ref{subsec:Projection-domain-shift})
and hubness problem (Sec. \ref{subsec:Hubness-Problem}).

\subsubsection{Projection Domain Shift Problems\label{subsec:Projection-domain-shift}}

\begin{figure}[t]
\centering{}\centering{}\includegraphics[scale=0.26]{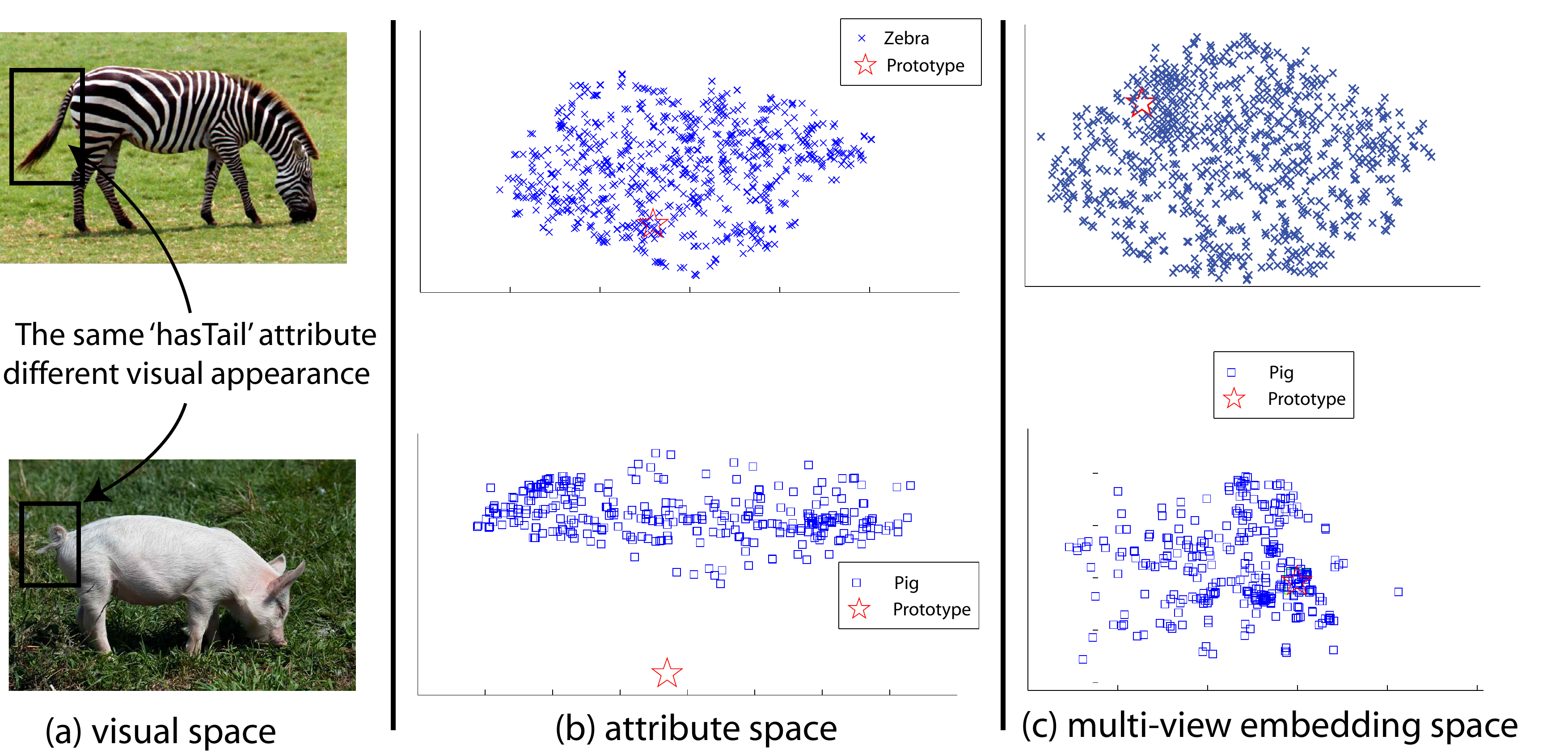}\caption{\label{fig:domain-shift:Low-level-feature-distribution}Illustrating
projection domain shift problem. Zero-shot prototypes are annotated
as red stars and predicted semantic attribute projections shown in
blue. Both Pig and Zebra share the same `hasTail' attribute yet with
very different visual appearance of `Tail'. The figure comes from
\cite{transductiveEmbeddingJournal}. }
\end{figure}

Projection domain shift problem in zero-shot recognition was first
identified by Fu \emph{et al.} \cite{transductiveEmbeddingJournal}.
This problem can be explained as follows: since the source and target
datasets have different classes, the underlying data distribution
of these classes may also differ. The projection functions learned
on the source dataset, from visual space to the embedding space, without
any adaptation to the target dataset, will cause an unknown shift/bias.
Figure \ref{fig:domain-shift:Low-level-feature-distribution} from
\cite{transductiveEmbeddingJournal} gives a more intuitive illustration
of this problem. It plots the 85D attribute space representation spanned
by feature projections which is learned from source data, and class
prototypes which are 85D binary attribute vectors. Zebra and Pig are
one of auxiliary and target classes respectively; and the same 'hasTail'
semantic attribute means very different visual appearance for Pig
and Zebra. In the attribute space, directly using the projection functions
learned from source datasets (\emph{e.g.}, Zebra) on the target datasets
(\emph{e.g.}, Pig) will lead to a large discrepancy between the class
prototype of the target class and the predicted semantic attribute
projections.

To alleviate this problem, the transductive learning based approaches
were proposed, to utilize the manifold information of the instances
from unseen classes \cite{transductiveEmbeddingJournal,Eylor_iccv2015,transferlearningNIPS,Li_CVPR2017,zsl_action_xu,yanweiembedding}.
Nevertheless, the transductive setting assumes that all the testing
data can be accessed at once, which obviously is invalid if the new
unseen classes appear dynamically and unavailable before learning
models. Thus inductive learning base approaches \cite{Eylor_iccv2015,synthesized_0shot,Jayaraman2014,semantic_graph,Li_CVPR2017}
have also been studied and these methods usually enforce other additional
constraints or information from the training data.

\subsubsection{Hubness problem\label{subsec:Hubness-Problem}}

The hubness problem is another interesting phenomenon that may be
observed in zero-shot recognition. Essentially, hubness problem can
be described as the presence of `universal' neighbors, or hubs, in
the space. Radovanovic \emph{et al. }\cite{marcobaronihubness} was
the first to study the hubness problem; in \cite{marcobaronihubness}
a hypothesis is made that hubness is an inherent property of data
distributions in the high dimensional vector space. Nevertheless,
Low \emph{et al.} \cite{Low2013} challenged this hypothesis and showed
the evidence that hubness is rather a boundary effect or, more generally,
an effect of a density gradient in the process of data generation.
Interestingly, their experiments showed that the hubness phenomenon
can also occur in low-dimensional data.

While causes for hubness are still under investigation, recent works
\cite{dinu2014improving,shigeto2015ridge} noticed that the regression
based zero-shot learning methods do suffer from this problem. To alleviate
this problem, Dinu \emph{et al.} \cite{dinu2014improving} utilized
the global distribution of feature instances of unseen data, \emph{i.e.},
in a transductive manner. In contrast, Yutaro \emph{et al.} \cite{shigeto2015ridge}
addressed this problem in an inductive way by embedding the class
prototypes into a visual feature space.

\section{Beyond Zero-shot Recognition\label{sec:Beyond-Zero-shot-Recognition}}

\subsection{Generalized Zero-shot Recognition and Open-set~Recognition}

In conventional supervised learning tasks, it is taken for granted
that the algorithms should take the form of ``closed set'' where
all testing classes should are known at training time. The zero-shot
recognition, in contrast, assumes that the source and target classes
cannot be mixed; and that the testing data only coming from the unseen
classes. This assumption, of course, greatly and unrealistically simplifies
the recognition tasks. To relax the settings of zero-shot recognition
and investigate recognition tasks in a more generic setting, there
are several tasks advocated beyond the conventional zero-shot recognition.
In particular, generalized zero-shot recognition \cite{wild_0shot}
and open set recognition tasks have been discussed recently \cite{Scheirer_2014_TPAMIb,Scheirer_2013_TPAMI,ssvoc_2016_CVPR,ssvoc_evl}.

The generalized zero-shot recognition proposed in \cite{wild_0shot}
broke the restricted nature of conventional zero-shot recognition
and also included the training classes among the testing data. Chao
\emph{et al.} \cite{wild_0shot} showed that it is nontrivial and
ineffective to directly extend the current zero-shot learning approaches
to solve the generalized zero-shot recognition. Such a generalized
setting, due to the more practical nature, is recommended as the evaluation
settings for zero-shot recognition tasks \cite{zsl_ugly}.

Open-set recognition, in contrast, has been developed independently
of zero-shot recognition. Initially, open set recognition aimed at
breaking the limitation of ``closed set'' recognition setup. Specifically,
the task of open set recognition tries to identify the class name
of an image from a very large set of classes, which includes but is
not limited to training classes. The open set recognition can be roughly
divided into two sub-groups.

\subsubsection{Conventional open set recognition}

First formulated in \cite{Bendale_2015_CVPR,Sattar_2015_CVPR,Scheirer_2013_TPAMI,Scheirer_2014_TPAMIb},
the conventional open set recognition only identifies whether the
testing images come from the training classes or some unseen classes.
This category of methods do not explicitly predict from which out
of unseen classes the testing instance, from the unseen classes, belongs
to. In such a setting, the conventional open set recognition is also
known as incremental learning \cite{gomes2008inc_dpmm,diehl2003inc_svm,iCaRL}.

\subsubsection{Generalized open set recognition}

The key difference from the conventional open set recognition is that
the generalized open set recognition also needs to explicitly predict
the semantic meaning (class) of testing instances even from the unseen
novel classes. This task was first defined and evaluated in \cite{ssvoc_2016_CVPR,ssvoc_evl}
on the tasks of object categorization. The generalized open set recognition
can be taken as a most general version of zero-shot recognition, where
the classifiers are trained from training instances of limited training
classes, whilst the learned classifiers are required to classify the
testing instances from a very large set of open vocabulary, say, 310
K class vocabulary in \cite{ssvoc_2016_CVPR,ssvoc_evl}. Conceptually
similar, there are vast variants of generalized open-set recognition
tasks which have been studied in other research community such as,
open-vocabulary object retrieval \cite{Guadarrama14:OOR,open_vocab_description},
open-world person re-identification \cite{open_world_1shot} or searching
targets \cite{Sattar_2015_CVPR}, open vocabulary scene parsing \cite{open_vocab_scen_parsing}.

\subsection{One-shot recognition}

A closely-related problem to zero-shot learning is one-shot or few-shot
learning problem \textendash{} instead of/apart from having only textual
description of the new classes, one-shot learning assumes that there
are one or few training samples for each class. Similar to zero-shot
recognition, one-shot recognition is inspired by fact that humans
are able to learn new object categories from one or very few examples
\cite{Jankowski,compositional_1shot}. Existing one-shot learning
approaches can be divided into two groups: the direct supervised learning
based approaches and the transfer learning based approaches.

\subsubsection{Direct Supervised Learning-based Approaches}

Early approaches do not assume that there exist a set of auxiliary
classes which are related and/or have ample training samples whereby
transferable knowledge can be extracted to compensate for the lack
of training samples. Instead, the target classes are used to trained
a standard classifier using supervised learning. The simplest method
is to employ nonparametric models such as kNN which are not restricted
by the number of training samples. However, without any learning,
the distance metric used for kNN is often inaccurate. To overcome
this problem, metric embedding can be learned and then used for kNN
classification \cite{NIPS2004_2566}. Other approaches attempt to
synthesize more training samples to augment the small training dataset
\cite{inverse_graphics,CAD_models,human_level_prob,compositional_1shot}.
However, without knowledge transfer from other classes, the performance
of direct supervised learning based approaches is typically weak.
Importantly, these models cannot meet the requirement of lifelong
learning, that is, when new unseen classes are added, the learned
classifier should still be able to recognize the seen existing classes.

\subsubsection{Transfer Learning-based One-shot Recognition}

This category of approaches follow a similar setting to zero-shot
learning, that is, they assume that an auxiliary set of training data
from different classes exist. They explore the paradigm of learning
to learn \cite{Thrun96learningto} or meta-learning \cite{JVilalta2002AIR}
and aim to transfer knowledge from the auxiliary dataset to the target
dataset with one or few examples per class. These approaches differ
in (i) what knowledge is transferred and (ii) how the knowledge is
represented. Specifically, the knowledge can be extracted and shared
in the form of model prior in a generative model \cite{feifei2003unsup_1s_objcat_learn,feifei2006one_shot,tommasi2009transfercat},
features \cite{bart2005cross_gen,hertz2016icml,Fleuret2005nips,amit2007icml,wolfc2005cvpr,torralba2005pami},
semantic attributes \cite{yanweiPAMIlatentattrib,lampert13AwAPAMI,transferlearningNIPS,rohrbach2010semantic_transfer},
or contextual information \cite{one_shot_TL_contexutal}. Many of
these approaches take a similar strategy as the existing zero-shot
learning approaches and transfer knowledge via a shared embedding
space. Embedding space can typically be formulated using neural networks
(\emph{e.g.}, siamese network \cite{Bromley1993ijcai,siamese_1shot}),
discriminative (\emph{e.g.}, Support Vector Regressors (SVR) \cite{farhadi2009attrib_describe,lampert13AwAPAMI,Kienzle2006icml}),
metric learning \cite{quattoni2008sparse_transfer,fink2005nips},
or kernel embedding \cite{wolf2009iccv,hertz2016icml} methods. Particularly,
one of most common embedding ways is semantic embedding which is normally
explored by projecting the visual features and semantic entities into
a common {\em new} space. Such projections can take various forms
with corresponding loss functions, such as SJE \cite{embedding_akata},
WSABIE \cite{Weston:2011:WSU:2283696.2283856}, ALE \cite{labelembeddingcvpr13},
DeViSE \cite{DeviseNIPS13}, and CCA \cite{yanweiembedding}.

More recently deep meta-learning has received increasing attention
for few-shot learning \cite{deep_1shot_recent,feedforward_1shot,siamese_1shot,video2vec_1shot,video_story_1shot,open_world_1shot,matchingnet_1shot,infield_1shot,compositional_1shot}.
Wang et al. \cite{yuxiong2016eccv,yuxiong2016nips} proposed the idea
of one-shot adaptation by automatically learning a generic, category
agnostic transformation from models learned from few samples to models
learned from large enough sample sets. A model-agnostic meta-learning
framework is proposed by Finn et al. \cite{pmlr-v70-finn17a} which
trains a deep model from the auxiliary dataset with the objective
that the learned model can be effectively updated/fine-tuned on the
new classes with one or few gradient steps. Note that similar to the
generalised zero-shot learning setting, recently the problem of adding
new classes to a deep neural network whilst keeping the ability to
recognise the old classes have been attempted \cite{rusu-progressive-2016}.
However, the problem of lifelong learning and progressively adding
new classes with few-shot remains an unsolved problem.

\section{Datasets in Zero-shot Recognition\label{sec:Datasets-in-Zero-shot}}

This section summarizes the datasets used for zero-shot recognition.
Recently with the increasing number of proposed zero-shot recognition
algorithms, Xian \emph{et al.} \cite{zsl_ugly} compared and analyzed
a significant number of the state-of-the-art methods in depth and
they defined a new benchmark by unifying both the evaluation protocols
and data splits. The details of these datasets are listed in Tab.
\ref{tab:Datasets-in-zero-shot}.

\subsection{Standard Datasets}

\subsubsection{Animal with Attribute (AwA) dataset \cite{lampert2009zeroshot_dat}}

AwA consists of the 50 Osher-son/Kemp animal category images collected
online. There are $30,475$ images with at least $92$ examples of
each class. Seven different feature types are provided: RGB color
histograms, SIFT~\cite{sift}, rgSIFT~\cite{colorSIFT2008CVPR},
PHOG~\cite{PHOG2007CVIR}, SURF~\cite{bay2008surf}, local self-similarity
histograms~\cite{selfsimilarity2007CVPR} and DeCaf~\cite{decaf2014ICML}.
The AwA dataset defines $50$ classes of animals, and $85$ associated
attributes (such as furry, and has claws). For the consistent evaluation
of attribute-based object classification methods, the AwA dataset
defined $10$ test classes: \emph{chimpanzee}, \emph{giant panda},
\emph{hippopotamus}, \emph{humpback whale}, \emph{leopard}, \emph{pig},
\emph{raccoon}, \emph{rat}, \emph{seal}. The $6,180$ images of those
classes are taken as the test data, whereas the $24,295$ images of
the remaining $40$ classes can be used for training. Since the images
in AwA are not available under a public license, Xian \emph{et al.}{}
\cite{zsl_ugly} introduced another new zero-shot learning dataset
\textendash{} Animals with Attributes 2 (AWA2) dataset with 37,322
publicly licensed and released images from the same 50 classes and
85 attributes as AwA. 

\subsubsection{aPascal-aYahoo dataset \cite{farhadi2009attrib_describe}}

aPascal-aYahoo has a 12,695-image subset of the PASCAL VOC 2008 data
set with $20$ object classes (aPascal); and 2,644 images that were
collected using the Yahoo image search engine (aYahoo) of $12$ object
classes. Each image in this data set has been annotated with 64 binary
attributes that characterize the visible objects.

\subsubsection{CUB-200-2011 dataset \cite{WahCUB_200_2011}}

CUB-200-2011 contains $11,788$ images of $200$ bird classes. This
is a more challenging dataset than AwA \textendash{} it is designed
for fine-grained recognition and has more classes but fewer images.
All images are annotated with bounding boxes, part locations, and
attribute labels. Images and annotations were filtered by multiple
users of Amazon Mechanical Turk. CUB-200-2011 is used as the benchmarks
dataset for multi-class categorization and part localization. Each
class is annotated with $312$ binary attributes derived from the
bird species ontology. A typical setting is to use $150$ classes
as auxiliary data, holding out $50$ as target data, which is the
setting adopted in Akata \emph{et al.} \cite{labelembeddingcvpr13}.

\subsubsection{Outdoor Scene Recognition (OSR) Dataset \cite{scene_OSR}}

OSR consists of $2,688$ images from $8$ categories and $6$ attributes
(`openness', `natrual', \emph{etc.}) and an average $426$ labelled
pairs for each attribute from $240$ training images. Graphs constructed
are thus extremely sparse. Pairwise attribute annotation was collected
by AMT (Kovashka \emph{et al. }\cite{whittlesearch}). Each pair was
labelled by $5$ workers to average the comparisons by majority voting.
Each image also belongs to a scene type.

\subsubsection{Public Figure Face Database (PubFig) \cite{kumar2009}}

PubFig is a large face dataset of 58,797 images of 200 people collected
from the internet. Parikh \emph{et al.} \cite{parikh2011relativeattrib}
selected a subset of PubFig consisting of $772$ images from $8$
people and $11$ attributes (`smiling', `round face', \emph{etc.}).
We annotate this subset as PubFig-sub. The pairwise attribute annotation
was collected by Amazon Mechanical Turk \cite{whittlesearch}. Each
pair was labelled by 5 workers. A total of 241 training images for
PubFig-sub respectively were labelled. The average number of compared
pairs per attribute were 418.

\subsubsection{SUN attribute dataset \cite{SUN_attrib}}

This is a subset of the SUN Database \cite{xiao2010sunscene} for
fine-grained scene categorization and it has $14,340$ images from
$717$ classes ($20$ images per class). Each image is annotated with
$102$ binary attributes that describe the scenes' material and surface
properties as well as lighting conditions, functions, affordances,
and general image layout.

\subsubsection{Unstructured Social Activity Attribute (USAA) dataset \cite{fu2012attribsocial}}

USAA is the first benchmark video attribute dataset for social activity
video classification and annotation. The ground-truth attributes are
annotated for $8$ semantic class videos of Columbia Consumer Video
(CCV) dataset~\cite{jiang2011consumervideo}, and select $100$ videos
per-class for training and testing respectively. These classes were
selected as the most complex social group activities. By referring
to the existing work on video ontology~\cite{Zha_ontology,jiang2011consumervideo},
the $69$ attributes can be divided into five broad classes: actions,
objects, scenes, sounds, and camera movement. Directly using the ground-truth
attributes as input to a SVM, the videos can come with $86.9\%$ classification
accuracy. This illustrates the challenge of USAA dataset: while the
attributes are informative, there is sufficient intra-class variability
in the attribute-space, and even perfect knowledge of the instance-level
attributes is also insufficient for perfect classification.

\subsubsection{ImageNet datasets \cite{rohrbach2010semantic_transfer,RohrbachCVPR12,ssvoc_2016_CVPR,synthesized_0shot}}

ImageNet has been used in several different papers with relatively
different settings. The original ImageNet dataset has been proposed
in \cite{deng2009imagenet}. The full set of ImageNet contains over
15 million labeled high-resolution images belonging to roughly 22,000
categories and labelled by human annotators using Amazon's Mechanical
Turk (AMT) crowd-sourcing tool. Starting in 2010, as part of the Pascal
Visual Object Challenge, an annual competition called the ImageNet
Large-Scale Visual Recognition Challenge (ILSVRC) has been held. ILSVRC
uses a subset of ImageNet with roughly 1,000 images in each of 1,000
categories. In \cite{transferlearningNIPS,RohrbachCVPR12}, Robhrbach
\emph{et al. }split the ILSVRC 2010 data into 800/200 classes for
source/target data. In \cite{ssvoc_2016_CVPR}, Fu \emph{et al. }employed
the training data of ILSVRC 2012 as the source data; and the testing
part of ILSVRC 2012 as well as the data of ILSVRC 2010 as the target
data. The full sized ImageNet data has been used in \cite{synthesized_0shot,DeviseNIPS13,ZSL_convex_optimization}.

\subsubsection{Oxford 102 Flower dataset \cite{oxford_flower}}

Oxford 102 is a collection of 102 groups of flowers each with 40 to
256 flower images, and total 8,189 images in total. The flowers were
chosen from the common flower species in the United Kingdom. Elhoseiny
\emph{et al.} \cite{Elhoseiny_2013_ICCV} generated textual descriptions
for each class of this dataset.

\begin{table*}
\centering{}%
\begin{tabular}{cccccc}
\hline 
 & Dataset  & \# instances  & \#classes  & \#attribute  & Annotation Level\tabularnewline
\hline 
\multirow{8}{*}{A } & AwA  & 30475  & 50  & 85  & per class\tabularnewline
 & aPascal-aYahoo  & 15339  & 32  & 64  & per image\tabularnewline
 & PubFig  & 58,797  & 200  & \textendash{}  & per image\tabularnewline
 & PubFig-sub  & 772  & 8  & 11  & per image pairs\tabularnewline
 & OSR  & 2688  & 8  & 6  & per image pairs\tabularnewline
 & ImageNet  & 15 million  & 22000  & \textendash{}  & per image\tabularnewline
 & ILSVRC 2010  & 1.2 million  & 1000  & \textendash{}  & per image\tabularnewline
 & ILSVRC 2012  & 1.2 million  & 1000  & \textendash{}  & per image\tabularnewline
\hline 
\hline 
\multirow{3}{*}{B } & Oxford 102 Flower  & 8189  & 102  & \textendash{}  & \textendash{}\tabularnewline
 & CUB-200-2011  & 11788  & 200  & 312  & per class\tabularnewline
 & SUN-attribute  & 14340  & 717  & 102  & per image\tabularnewline
\hline 
\hline 
\multirow{4}{*}{C } & USAA  & 1600  & 8  & 69  & per video\tabularnewline
 & UCF101  & 13320  & 101  & \textendash{}  & per video\tabularnewline
 & ActivityNet  & 27801  & 203  & \textendash{}  & per video\tabularnewline
 & FCVID  & 91223  & 239  & \textendash{}  & per video\tabularnewline
\hline 
\end{tabular}\caption{\label{tab:Datasets-in-zero-shot}Datasets in zero-shot recognition.
The datasets are divided into three groups: general image classification
(A), fine-grained image classification (B) and video classification
datasets (C).}
\end{table*}

\subsubsection{UCF101 dataset \cite{ucf101}}

UCF101 is another popular benchmark for human action recognition in
videos, which consists of $13,320$ video clips (27 hours in total)
with 101 annotated classes. More recently, the THUMOS-2014 Action
Recognition Challenge \cite{THUMOS} created a benchmark by extending
upon the UCF-101 dataset (used as the training set). Additional videos
were collected from the Internet, including $2,500$ background videos,
$1,000$ validation and $1,574$ test videos.

\subsubsection{Fudan-Columbia Video Dataset (FCVID) \cite{fcvid_2017}}

FCVID contains $91,223$ web videos annotated manually into $239$
categories. Categories cover a wide range of topics (not only activities),
such as social events (\emph{e.g.}, tailgate party), procedural events
(\emph{e.g.}, making cake), object appearances (\emph{e.g.}, panda)
and scenic videos (\emph{e.g.}, beach). Standard split consists of
$45,611$ videos for training and $45,612$ videos for testing.

\subsubsection{ActivityNet dataset \cite{activitynet}}

ActivityNet is another large-scale video dataset for human activity
recognition and understanding and released in 2015. It consisted of
27,801 video clips annotated into 203 activity classes, totaling 849
hours of video. Comparing with existing dataset, ActivityNet has more
fine-grained action categories (\emph{e.g.}, ``drinking beer\textquotedblright{}
and ``drinking coffee\textquotedblright ). ActivityNet had the settings
of both trimmed and untrimmed videos of its classes.

\subsection{Discussion of Datasets.}

In Tab. \ref{tab:Datasets-in-zero-shot}, we roughly divide all the
datasets into three groups: general image classification, fine-grained
image classification and video classification datasets. These datasets
have been employed widely as the benchmark datasets in many previous
works. However, we believe that when making a comparison with the
other existing methods on these datasets, there are several issues
that should be discussed.

\subsubsection{Features}

With the renaissance of deep convolutional neural networks, deep features
of images/videos have been used for zero-shot recognition. Note that
different types of deep features (\emph{e.g.}, Overfeat \cite{overfeat},
VGG-19\cite{returnDevil2014BMVC}, or ResNet \cite{he2015deep}) have
varying level of semantic abstraction and representation ability;
and even the same type of deep features, if fine-tuned on different
dataset and with slightly different parameters, will also have different
representative ability. Thus it should be obvious, without using the
same type of features, it is not possible to conduct a fair comparisons
among different methods and draw any meaningful conclusion. Importantly
it is possible that the improved performance of one zero shot recognition
could be largely attributed to the better deep features used.

\subsubsection{Auxiliary data}

As mentioned, zero-shot recognition can be formulated in a transfer
learning setting. The size and quality of auxiliary data can be very
important for the overall performance of zero-shot recognition. Note
that these auxiliary data do not only include the auxiliary source
image/video dataset, but also refer to the data to extract/train the
concept ontology, or semantic word vectors. For example, the semantic
word vectors trained on large-scale linguistic articles, in general,
are better semantically distributed than those trained on small sized
linguistic corpus. Similarly, GloVe \cite{GloVec} is reported to
be better than the skip-gram and CBOW models \cite{distributedword2vec2013NIPS}.
Therefore, to make a fair comparison with existing works, another
important factor is to use the same set of auxiliary data.

\subsubsection{Evaluation}

For many datasets, there is no agreed source/target splits for zero-shot
evaluation. Xian \emph{et al.} \cite{zsl_ugly} suggested a new benchmark
by unifying both the evaluation protocols and data splits.

\section{Future Research Directions\label{sec:Future-Research-Directions}}

\subsubsection{More Generalized and Realistic Setting}

From the detailed review of existing zero-shot learning methods, it
is clear that overall the existing efforts have been focused on a
rather restrictive and impractical setting: classification is required
for new object classes only and the new unseen classes, though having
no training sample present, are assumed to be known. In reality, one
wants to progressively add new classes to the existing classes. Importantly,
this needs to be achieved without jeopardizing the ability of the
model to recognize existing seen classes. Furthermore, we cannot assume
that the new samples will only come from a set of known unseen classes.
Rather, they can only be assumed to belong to either existing seen
classes, known unseen classes, or unknown unseen classes. We therefore
foresee a more generalized setting will be adopted by the future zero-shot
learning work.

\subsubsection{Combining Zero-shot with Few-shot Learning}

As mentioned earlier, the problems of zero-shot and few-shot learning
are closely related and as a result, many existing methods use the
same or similar models. However, it is somewhat surprising to note
that no serious efforts have been taken to address these two problems
jointly. In particular, zero-shot learning would typically not consider
the possibility of having few training samples, while few-shot learning
ignores the fact that the textual description/human knowledge about
the new class is always there to be exploited. A few existing zero-shot
learning methods \cite{ssvoc_2016_CVPR,yanweiPAMIlatentattrib,latent_0shot,deep_0shot_cvpr}
have included few-shot learning experiments. However, they typically
use a naive \emph{kNN} approach, that is, each class prototype is
treated as a training sample and together with the k-shot, this becomes
a k+1-shot recognition problem. However, as shown by existing zero-shot
learning methods \cite{transductiveEmbeddingJournal}, the prototype
is worth far more that one training sample; it thus should be treated
differently. We thus expect a future direction on extending the existing
few-shot learning methods by incorporating the prototype as a `super'-shot
to improve the model learning.

\subsubsection{Beyond object categories}

So far the current zero-shot learning efforts are limited to recognizing
object categories. However, visual concepts can have far complicated
relationships than object categories. In particular beyond objects/nouns,
attributes/adjectives are important visual concepts. When combined
with objects, the same attribute often has different meaning, \emph{e.g.},
the concept of `yellow' in yellow face and a yellow banana clearly
differs. Zero-shot learning attributes with associated objects is
thus an interesting future research direction.

\subsubsection{Curriculum learning}

In a lifelong learning setting, a model will incrementally learn to
recognise new classes whilst keep the capacity for existing classes.
A related problem is thus how to select the more suitable new classes
to learn given the existing classes. It has been shown that \cite{iCaRL,lifelong_iid,curriculum_learning}
the sequence of adding different classes have a clear impact on the
model performance. It is therefore useful to investigate how to incorporate
the curriculum learning principles in designing a zero-shot learning
strategy.

\section{Conclusion\label{sec:Conclusion}}

In this paper, we have reviewed the recent advances in zero shot recognition.
Firstly different types of semantic representations are examined and
compared; the models used in zero shot learning have also been investigated.
Next, beyond zero shot recognition, one-shot and open set recognition
are identified as two very important related topics and thus reviewed.
Finally, the common used datasets in zero-shot recognition have been
reviewed with a number of issues in existing evaluations of zero-shot
recognition methods discussed. We also point out a number of research
direction which we believe will the focus of the future zero-shot
recognition studies.

\vspace{0.1in}
\noindent \textbf{Acknowledgments.}
This work is supported in part by two grants from NSF China ($\#61702108$, $\#61622204$, $\#61572134$),  and an European FP7 project (PIRSESGA-$2013-612652$). Yanwei Fu is supported by The Program for Professor of Special Appointment (Eastern Scholar) at Shanghai Institutions of Higher Learning.
\noindent  \bibliographystyle{IEEEtran}
\bibliography{egbib}

\begin{thebibliography}{100}
\providecommand{\url}[1]{#1}
\csname url@samestyle\endcsname
\providecommand{\newblock}{\relax}
\providecommand{\bibinfo}[2]{#2}
\providecommand{\BIBentrySTDinterwordspacing}{\spaceskip=0pt\relax}
\providecommand{\BIBentryALTinterwordstretchfactor}{4}
\providecommand{\BIBentryALTinterwordspacing}{\spaceskip=\fontdimen2\font plus
\BIBentryALTinterwordstretchfactor\fontdimen3\font minus
  \fontdimen4\font\relax}
\providecommand{\BIBforeignlanguage}[2]{{%
\expandafter\ifx\csname l@#1\endcsname\relax
\typeout{** WARNING: IEEEtran.bst: No hyphenation pattern has been}%
\typeout{** loaded for the language `#1'. Using the pattern for}%
\typeout{** the default language instead.}%
\else
\language=\csname l@#1\endcsname
\fi
#2}}
\providecommand{\BIBdecl}{\relax}
\BIBdecl

\bibitem{object_cat_1987}
I.~Biederman, ``Recognition by components - a theory of human image
  understanding,'' \emph{Psychological Review}, 1987.

\bibitem{chen_iccv13}
X.~Chen, A.~Shrivastava, and A.~Gupta, ``{NEIL}: {E}xtracting {V}isual
  {K}nowledge from {W}eb {D}ata,'' in \emph{IEEE International Conference on
  Computer Vision}, 2013.

\bibitem{lifelonglearning}
A.~Pentina and C.~H. Lampert, ``A {PAC}-bayesian bound for lifelong learning,''
  in \emph{International Conference on Machine Learning}, 2014.

\bibitem{Tom1995lifelong}
S.~Thrun and T.~M. Mitchell, ``Lifelong robot learning,'' \emph{Robotics and
  Autonomous Systems}, 1995.

\bibitem{pan2009transfer_survey}
S.~J. Pan and Q.~Yang, ``A survey on transfer learning,'' \emph{IEEE
  Transactions on Data and Knowledge Engineering}, vol.~22, no.~10, pp.
  1345--1359, 2010.

\bibitem{visual_domain_adapt}
V.~Patel, R.~Gopalan, R.~Li, and R.~Chellappa, ``Visual domain adaptation: A
  survey of recent advances,'' \emph{IEEE Signal Processing Magazine (SPM)},
  2015.

\bibitem{palatucci2009zero_shot}
M.~Palatucci, G.~Hinton, D.~Pomerleau, and T.~M. Mitchell, ``Zero-shot learning
  with semantic output codes,'' in \emph{NIPS}, 2009.

\bibitem{kumar2009}
N.~Kumar, A.~C. Berg, P.~N. Belhumeur, and S.~K. Nayar, ``Attribute and simile
  classifiers for face verification,'' in \emph{ICCV}, 2009.

\bibitem{lampert13AwAPAMI}
C.~H. Lampert, H.~Nickisch, and S.~Harmeling, ``Attribute-based classification
  for zero-shot visual object categorization,'' \emph{IEEE TPAMI}, 2013.

\bibitem{emotion_0shot}
B.~Xu, Y.~Fu, Y.-G. Jiang, B.~Li, and L.~Sigal, ``Video emotion recognition
  with transferred deep feature encodings,'' in \emph{ICMR}, 2016.

\bibitem{fu2012attribsocial}
Y.~Fu, T.~Hospedales, T.~Xiang, and S.~Gong, ``Attribute learning for
  understanding unstructured social activity,'' in \emph{ECCV}, 2012.

\bibitem{liu2011action_attrib}
J.~Liu, B.~Kuipers, and S.~Savarese, ``Recognizing human actions by
  attributes,'' in \emph{IEEE Conference on Computer Vision and Pattern
  Recognition}, 2011.

\bibitem{yanweiPAMIlatentattrib}
Y.~Fu, T.~M. Hospedales, T.~Xiang, and S.~Gong, ``Learning multi-modal latent
  attributes,'' \emph{IEEE TPAMI}, 2013.

\bibitem{Blitzer_zero-shotdomain}
J.~Blitzer, D.~P. Foster, and S.~M. Kakade, ``Zero-shot domain adaptation: A
  multi-view approach,'' TTI-TR-2009-1, Tech. Rep., 2009.

\bibitem{RichardNIPS13}
R.~Socher, M.~Ganjoo, H.~Sridhar, O.~Bastani, C.~D. Manning, and A.~Y. Ng,
  ``Zero-shot learning through cross-modal transfer,'' in \emph{NIPS}, 2013.

\bibitem{Thrun96learningto}
S.~Thrun, \emph{Learning To Learn: Introduction}.\hskip 1em plus 0.5em minus
  0.4em\relax Kluwer Academic Publishers, 1996.

\bibitem{farhadi2009attrib_describe}
A.~Farhadi, I.~Endres, D.~Hoiem, and D.~Forsyth, ``Describing objects by their
  attributes,'' in \emph{CVPR}, 2009.

\bibitem{wang2011clothesattrib}
\BIBentryALTinterwordspacing
X.~Wang and T.~Zhang, ``Clothes search in consumer photos via color matching
  and attribute learning,'' in \emph{ACM International Conference on
  Multimedia}, 2011. [Online]. Available:
  \url{http://doi.acm.org/10.1145/2072298.2072013}
\BIBentrySTDinterwordspacing

\bibitem{torralba2011app_share}
R.~Salakhutdinov, A.~Torralba, and J.~Tenenbaum, ``Learning to share visual
  appearance for multiclass object detection,'' in \emph{IEEE Conference on
  Computer Vision and Pattern Recognition}, 2011.

\bibitem{hwang2011obj_attrib}
S.~J. Hwang, F.~Sha, and K.~Grauman, ``Sharing features between objects and
  their attributes,'' in \emph{IEEE Conference on Computer Vision and Pattern
  Recognition}, 2011.

\bibitem{parikh2011relativeattrib}
D.~Parikh and K.~Grauman, ``Relative attributes,'' in \emph{ICCV}, 2011.

\bibitem{whittlesearch}
A.~Kovashka, D.~Parikh, and K.~Grauman, ``{WhittleSearch}: Image search with
  relative attribute feedback,'' in \emph{IEEE Conference on Computer Vision
  and Pattern Recognition}, 2012.

\bibitem{attrbDiscovery12ECCV}
T.~L. Berg, A.~C. Berg, and J.~Shih, ``Automatic attribute discovery and
  characterization from noisy web data,'' in \emph{European Conference on
  Computer Vision}, 2010.

\bibitem{robust_0shot}
Y.~Fu, T.~M. Hospedales, J.~Xiong, T.~Xiang, S.~Gong, Y.~Yao, and Y.~Wang,
  ``Robust estimation of subjective visual properties from crowdsourced
  pairwise labels,'' \emph{IEEE TPAMI}, 2016.

\bibitem{imginterestingnessICCV2013}
M.~Gygli, H.~Grabner, H.~Riemenschneider, F.~Nater, and L.~V. Gool, ``The
  interestingness of images,'' in \emph{IEEE International Conference on
  Computer Vision}, 2013.

\bibitem{yugangVideoInteresting2013}
Y.-G. Jiang, YanranWang, R.~Feng, X.~Xue, Y.~Zheng, and H.~Yang,
  ``Understanding and predicting interestingness of videos,'' in \emph{AAAI
  Conference on Artificial Intelligence}, 2013.

\bibitem{Isola2011NIPS}
P.~Isola, D.~Parikh, A.~Torralba, and A.~Oliva, ``Understanding the intrinsic
  memorability of images,'' in \emph{Neural Information Processing Systems},
  2011.

\bibitem{Isola2011cvpr}
P.~Isola, J.~Xiao, A.~Torralba, and A.~Oliva, ``What makes an image
  memorable?'' in \emph{IEEE Conference on Computer Vision and Pattern
  Recognition}, 2011.

\bibitem{Dhar2011cvpr}
S.~Dhar, V.~Ordonez, and T.~L. Berg, ``High level describable attributes for
  predicting aesthetics and interestingness,'' in \emph{IEEE Conference on
  Computer Vision and Pattern Recognition}, 2011.

\bibitem{fu2010ageSurvey}
Y.~Fu, G.~Guo, and T.~Huang, ``Age synthesis and estimation via faces: A
  survey,'' \emph{IEEE Transactions on Pattern Analysis and Machine
  Intelligence}, 2010.

\bibitem{crowdcountingKE}
K.~Chen, S.~Gong, T.~Xiang, and C.~C. Loy, ``Cumulative attribute space for age
  and crowd density estimation,'' in \emph{IEEE Conference on Computer Vision
  and Pattern Recognition}, 2013.

\bibitem{lampert2009zeroshot_dat}
C.~H. Lampert, H.~Nickisch, and S.~Harmeling, ``Learning to detect unseen
  object classes by between-class attribute transfer,'' in \emph{CVPR}, 2009.

\bibitem{moon_attrb}
E.~M. Rudd, M.~Gunther, and T.~E. Boult, ``Moon:a mixed objective optimization
  network for the recognition of facial attributes,'' in \emph{ECCV}, 2016.

\bibitem{rudd2016moon}
E.~Rudd, M.~G{\"u}nther, and T.~Boult, ``Moon: A mixed objective optimization
  network for the recognition of facial attributes,'' \emph{arXiv preprint
  arXiv:1603.07027}, 2016.

\bibitem{wang2016walk}
J.~Wang, Y.~Cheng, and R.~S. Feris, ``Walk and learn: Facial attribute
  representation learning from egocentric video and contextual data,'' in
  \emph{CVPR}, 2016.

\bibitem{datta2011face_attrib}
A.~Datta, R.~Feris, and D.~Vaquero, ``Hierarchical ranking of facial
  attributes,'' in \emph{IEEE International Conference on Automatic Face \&
  Gesture Recognition}, 2011.

\bibitem{ehrlich2016facial}
M.~Ehrlich, T.~J. Shields, T.~Almaev, and M.~R. Amer, ``Facial attributes
  classification using multi-task representation learning,'' in
  \emph{Proceedings of the IEEE Conference on Computer Vision and Pattern
  Recognition Workshops}, 2016, pp. 47--55.

\bibitem{survey_of_face}
R.~Jafri and H.~R. Arabnia, ``A survey of face recognition techniques,'' in
  \emph{Journal of Information Processing Systems}, 2009.

\bibitem{facial_attrb_icmr}
Z.~Wang, K.~He, Y.~Fu, R.~Feng, Y.-G. Jiang, and X.~Xue, ``Multi-task deep
  neural network for joint face recognition and facial attribute prediction,''
  in \emph{ACM ICMR}, 2017.

\bibitem{multi_task_attrib}
A.~Argyriou, T.~Evgeniou, and M.~Pontil, ``Convex multi-task feature
  learning,'' in \emph{ACM ICMR}, 2007.

\bibitem{3D_shape_attribute}
D.~F. Fouhey, A.~Gupta, and A.~Zisserman, ``Understanding higher-order shape
  via 3d shape attributes,'' in \emph{IEEE TPAMI}, 2017.

\bibitem{vaquero2009attrib_surveil}
D.~Vaquero, R.~Feris, D.~Tran, L.~Brown, A.~Hampapur, and M.~Turk,
  ``Attribute-based people search in surveillance environments,'' in \emph{IEEE
  Workshop on Applications of Computer Vision (WACV)}, dec. 2009, pp. 1 --8.

\bibitem{wang2009attrib_class_sal}
G.~Wang and D.~Forsyth, ``Joint learning of visual attributes, object classes
  and visual saliency,'' in \emph{IEEE International Conference on Computer
  Vision}, 2009, pp. 537--544.

\bibitem{ferrari2007attrib_learn}
V.~Ferrari and A.~Zisserman, ``Learning visual attributes,'' in \emph{Neural
  Information Processing Systems}, Dec. 2007.

\bibitem{ShrivastavaECCV12}
A.~Shrivastava, S.~Singh, and A.~Gupta, ``Constrained semi-supervised learning
  via attributes and comparative attributes,'' in \emph{European Conference on
  Computer Vision}, 2012.

\bibitem{BiswasCVPR13}
A.~Biswas and D.~Parikh, ``Simultaneous active learning of classifiers and
  attributes via relative feedback,'' in \emph{IEEE Conference on Computer
  Vision and Pattern Recognition}, 2013.

\bibitem{attr_clas_feedback}
A.~Parkash and D.~Parikh, ``Attributes for classifier feedback,'' in
  \emph{European Conference on Computer Vision}, 2012.

\bibitem{relative_ranking_eccv16}
K.~K. Singh and Y.~J. Lee, ``End-to-end localization and ranking for relative
  attributes,'' in \emph{ECCV}, 2016.

\bibitem{jaderberg2015spatial}
M.~Jaderberg, K.~Simonyan, A.~Zisserman \emph{et~al.}, ``Spatial transformer
  networks,'' in \emph{Advances in Neural Information Processing Systems},
  2015, pp. 2017--2025.

\bibitem{parikh2011nameable_attribs}
D.~Parikh and K.~Grauman, ``Interactively building a discriminative vocabulary
  of nameable attributes,'' in \emph{IEEE Conference on Computer Vision and
  Pattern Recognition}, 2011.

\bibitem{tang2009concepts_from_noisytags}
\BIBentryALTinterwordspacing
J.~Tang, S.~Yan, R.~Hong, G.-J. Qi, and T.-S. Chua, ``Inferring semantic
  concepts from community-contributed images and noisy tags,'' in \emph{ACM
  International Conference on Multimedia}, 2009. [Online]. Available:
  \url{http://doi.acm.org/10.1145/1631272.1631305}
\BIBentrySTDinterwordspacing

\bibitem{video_story_1shot}
A.~Habibian, T.~Mensink, and C.~Snoek, ``Videostory: A new multimedia embedding
  for few-example recognition and translation of events,'' in \emph{ACM MM},
  2014.

\bibitem{hauptmann2007semanticGapRetr}
A.~Hauptmann, R.~Yan, W.-H. Lin, M.~Christel, and H.~Wactlar, ``Can high-level
  concepts fill the semantic gap in video retrieval? a case study with
  broadcast news,'' \emph{IEEE Transactions on Multimedia}, vol.~9, no.~5, pp.
  958 --966, aug. 2007.

\bibitem{snoek2007semantic_retrieval}
C.~G.~M. Snoek, B.~Huurnink, L.~Hollink, M.~de~Rijke, G.~Schreiber, and
  M.~Worring, ``Adding semantics to detectors for video retrieval,'' \emph{IEEE
  Transactions on Multimedia}, vol.~9, pp. 975--986, 2007.

\bibitem{toderici2010youtube_tag}
G.~Toderici, H.~Aradhye, M.~Pasca, L.~Sbaiz, and J.~Yagnik, ``Finding meaning
  on youtube: Tag recommendation and category discovery,'' in \emph{IEEE
  Conference on Computer Vision and Pattern Recognition}, 2010, pp. 3447--3454.

\bibitem{zuxuan_2016_CVPR}
Z.~Wu, Y.~Fu, Y.-G. Jiang, and L.~Sigal, ``Harnessing object and scene
  semantics for large-scale video understanding,'' in \emph{CVPR}, 2016.

\bibitem{obj2action}
M.~Jain, J.~C. van Gemert, T.~Mensink, and C.~G.~M. Snoek, ``Objects2action:
  Classifying and localizing actions without any video example,'' in
  \emph{ICCV}, 2015.

\bibitem{tang2009annotation}
J.~Tang, X.-S. Hua, M.~Wang, Z.~Gu, G.-J. Qi, and X.~Wu, ``Correlative linear
  neighborhood propagation for video annotation,'' \emph{IEEE Transactions on
  Systems, Man, and Cybernetics, Part B}, vol.~39, no.~2, pp. 409--416, 2009.

\bibitem{qi2007corr_mlab}
\BIBentryALTinterwordspacing
G.-J. Qi, X.-S. Hua, Y.~Rui, J.~Tang, T.~Mei, and H.-J. Zhang, ``Correlative
  multi-label video annotation,'' in \emph{ACM International Conference on
  Multimedia}, 2007. [Online]. Available:
  \url{http://doi.acm.org/10.1145/1291233.1291245}
\BIBentrySTDinterwordspacing

\bibitem{fergus2010label_share}
R.~Fergus, H.~Bernal, Y.~Weiss, and A.~Torralba, ``Semantic label sharing for
  learning with many categories,'' in \emph{European Conference on Computer
  Vision}, 2010.

\bibitem{RohrbachCVPR12}
M.~Rohrbach, M.~Stark, and B.~Schiele, ``Evaluating knowledge transfer and
  zero-shot learning in a large-scale setting,'' in \emph{CVPR}, 2012.

\bibitem{rohrbach2010semantic_transfer}
M.~Rohrbach, M.~Stark, G.~Szarvas, I.~Gurevych, and B.~Schiele, ``What helps
  where -- and why? semantic relatedness for knowledge transfer,'' in
  \emph{CVPR}, 2010.

\bibitem{costa_mlzsl}
T.~Mensink, E.~Gavves, and C.~G. Snoek, ``Costa: Co-occurrence statistics for
  zero-shot classification,'' in \emph{IEEE Conference on Computer Vision and
  Pattern Recognition}, 2014.

\bibitem{recog_action}
C.~Gan, Y.~Yang, L.~Zhu, and Y.~Zhuang, ``Recognizing an action using its name:
  A knowledge-based approach,'' \emph{IJCV}, 2016.

\bibitem{concept_not_alone}
C.~Gan, M.~Lin, Y.~Yang, G.~de~Melo, and A.~G. Hauptmann, ``Concepts not alone:
  Exploring pairwise relationships for zero-shot video activity recognition,''
  in \emph{AAAI}, 2016.

\bibitem{ZSL_convex_optimization}
M.~Norouzi, T.~Mikolov, S.~Bengio, Y.~Singer, J.~Shlens, A.~Frome, G.~S.
  Corrado, and J.~Dean, ``Zero-shot learning by convex combination of semantic
  embeddings,'' \emph{ICLR}, 2014.

\bibitem{zhang2016zero}
Z.~Zhang and V.~Saligrama, ``Zero-shot learning via joint latent similarity
  embedding,'' in \emph{CVPR}, 2016.

\bibitem{zhang2016zeroshot}
------, ``Zero-shot recognition via structured prediction,'' in \emph{ECCV},
  2016.

\bibitem{yanweiBMVC}
Y.~Fu, Y.~Yang, T.~Hospedales, T.~Xiang, and S.~Gong, ``Transductive
  multi-label zero-shot learning,'' in \emph{British Machine Vision
  Conference}, 2014.

\bibitem{DeviseNIPS13}
A.~Frome, G.~S. Corrado, J.~Shlens, S.~Bengio, J.~Dean, M.~Ranzato, and
  T.~Mikolov, ``{DeViSE}: A deep visual-semantic embedding model,'' in
  \emph{NIPS}, 2013.

\bibitem{huang2012ACL}
E.~H. Huang, R.~Socher, C.~D. Manning, and A.~Y. Ng, ``Improving word
  representations via global context and multiple word prototypes,'' in
  \emph{Association for Computational Linguistics 2012 Conference}, 2012.

\bibitem{TAC_0shot}
B.~Xu, Y.~Fu, Y.-G. Jiang, B.~Li, and L.~Sigal, ``Heterogeneous knowledge
  transfer in video emotion recognition, attribution and summarization,''
  \emph{IEEE TAC}, 2016.

\bibitem{amazon_mechanical}
A.~Sorokin and D.~Forsyth, ``Utility data annotation with amazon mechanical
  turk,'' in \emph{IEEE Conference on Computer Vision and Pattern Recognition
  Workshops}, 2008.

\bibitem{latent_semantic_attrb}
J.~Qin, Y.~Wang, L.~Liu, J.~Chen, and L.~Shao, ``Beyond semantic attributes:
  Discrete latent attributes learning for zero-shot recognition,'' \emph{IEEE
  Signal Processing Letters}, vol.~23, no.~11, pp. 1667--1671, 2016.

\bibitem{change_aaai}
X.~Chang, Y.~Yang, G.~Long, C.~Zhang, and A.~Hauptmann, ``Dynamic concept
  composition for zero-example event detection,'' in \emph{AAAI}, 2016.

\bibitem{change_ijcai}
X.~Chang, Y.~Yang, A.~Hauptmann, E.~P. Xing, and Y.~Yu, ``Semantic concept
  discovery for large-scale zero-shot event detection,'' in \emph{IJCAI}, 2015.

\bibitem{zero_shot_action_cvpr2017}
J.~Qin, L.~Liu, L.~Shao, F.~Shen, B.~Ni, J.~Chen, , and YunhongWang,
  ``Zero-shot action recognition with error-correcting output codes,'' in
  \emph{CVPR}, 2017.

\bibitem{yang2011tag_tagging}
K.~Yang, X.-S. Hua, M.~Wang, and H.-J. Zhang, ``Tag tagging: Towards more
  descriptive keywords of image content,'' \emph{IEEE Transactions on
  Multimedia}, vol.~13, pp. 662 --673, 2011.

\bibitem{hospedales2011video_tags}
T.~Hospedales, S.~Gong, and T.~Xiang, ``Learning tags from unsegmented videos
  of multiple human actions,'' in \emph{International Conference on Data
  Mining}, 2011.

\bibitem{Aradhye2009}
H.~Aradhye, G.~Toderici, and J.~Yagnik, ``Video2text: Learning to annotate
  video content,'' in \emph{Proc. IEEE Int. Conf. Data Mining Workshops ICDMW
  '09}, 2009, pp. 144--151.

\bibitem{yang2011disc_subtag}
W.~Yang and G.~Toderici, ``Discriminative tag learning on youtube videos with
  latent sub-tags,'' in \emph{IEEE Conference on Computer Vision and Pattern
  Recognition}, 2011.

\bibitem{multimodal_0shot}
S.~Wu, F.~Luisier, and S.~Bondugula, ``Zero-shot event detection using
  multi-modal fusion of weakly supervised concepts,'' in \emph{CVPR}, 2016.

\bibitem{Elhoseiny_2013_ICCV}
M.~Elhoseiny, B.~Saleh, and A.~Elgammal, ``Write a classifier: Zero-shot
  learning using purely textual descriptions,'' in \emph{IEEE International
  Conference on Computer Vision}, December 2013.

\bibitem{deep_0shot}
J.~L. Ba, K.~Swersky, S.~Fidler, and R.~Salakhutdinov, ``Predicting deep
  zero-shot convolutional neural networks using textual descriptions,'' in
  \emph{ICCV}, 2015.

\bibitem{deep_0shot_cvpr}
S.~Reed, Z.~Akata, B.~Schiele, and H.~Lee., ``Learning deep representations of
  fine-grained visual descriptions,'' in \emph{CVPR}, 2016.

\bibitem{WordNet_1995Miller}
G.~A. Miller, ``Wordnet: A lexical database for english,'' \emph{Commun. ACM},
  vol.~38, no.~11, pp. 39--41, Nov. 1995.

\bibitem{wordvectorICLR}
T.~Mikolov, K.~Chen, G.~Corrado, and J.~Dean, ``Efficient estimation of word
  representation in vector space,'' in \emph{Proceedings of Workshop at
  International Conference on Learning Representations}, 2013.

\bibitem{distributedword2vec2013NIPS}
T.~Mikolov, I.~Sutskever, K.~Chen, G.~Corrado, and J.~Dean, ``Distributed
  representations of words and phrases and their compositionality,'' in
  \emph{Neural Information Processing Systems}, 2013.

\bibitem{Harris1981}
Z.~S. Harris, \emph{Distributional Structure}.\hskip 1em plus 0.5em minus
  0.4em\relax Dordrecht: Springer Netherlands, 1981, pp. 3--22.

\bibitem{zero_data_AAAI2008}
H.~Larochelle, D.~Erhan, and Y.~Bengio, ``Zero-data learning of new tasks,'' in
  \emph{AAAI}, 2008.

\bibitem{yu2010attributetransfer}
X.~Yu and Y.~Aloimonos, ``Attribute-based transfer learning for object
  categorization with zero/one training example,'' in \emph{European Conference
  on Computer Vision}, 2010.

\bibitem{Jayaraman2014}
D.~Jayaraman and K.~Grauman, ``Zero shot recognition with unreliable
  attributes,'' in \emph{NIPS}, 2014.

\bibitem{labelembeddingcvpr13}
Z.~Akata, F.~Perronnin, Z.~Harchaoui, and C.~Schmid, ``Label-embedding for
  attribute-based classification,'' in \emph{CVPR}, 2013.

\bibitem{WASABIE2010}
J.~Weston, S.~Bengio, and N.~Usunier, ``Large scale image annotation: learning
  to rank with joint word-image embeddings,'' \emph{Machine Learning}, 2010.

\bibitem{LiECCV2014}
Z.~Li, E.~Gavves, T.~E.~J. Mensink, and C.~G.~M. Snoek, ``Attributes make sense
  on segmented objects,'' in \emph{European Conference on Computer Vision},
  2014.

\bibitem{max_margin_zsl_2015}
X.~Li and Y.~Guo, ``Max-margin zero-shot learning for multiclass
  classification,'' in \emph{AISTATS}, 2015.

\bibitem{sslzsl_0shot}
X.~Li, Y.~Guo, and D.~Schuurmans, ``Semi-supervised zero-shot classification
  with label representation learning,'' in \emph{ICCV}, 2015.

\bibitem{decorrelated_cvpr14}
D.~Jayaraman, F.~Sha, and K.~Grauman, ``Decorrelating semantic visual
  attributes by resisting the urge to share,'' in \emph{CVPR}, 2014.

\bibitem{unified_model}
S.~J. Hwang and L.~Sigal, ``A unified semantic embedding: relating taxonomies
  and attributes,'' in \emph{NIPS}, 2014.

\bibitem{embedding_akata}
Z.~Akata, S.~Reed, D.~Walter, H.~Lee, and B.~Schiele, ``Evaluation of output
  embeddings for fine-grained image classification,'' in \emph{CVPR}, 2015.

\bibitem{romera2015embarrassingly}
B.~Romera-Paredes and P.~Torr, ``An embarrassingly simple approach to zero-shot
  learning,'' in \emph{ICML}, 2015.

\bibitem{yanweiembedding}
Y.~Fu, T.~M. Hospedales, T.~Xiang, Z.~Fu, and S.~Gong, ``Transductive
  multi-view embedding for zero-shot recognition and annotation,'' in
  \emph{ECCV}, 2014.

\bibitem{yang2014unified}
Y.~Yang and T.~M. Hospedales, ``A unified perspective on multi-domain and
  multi-task learning,'' in \emph{ICLR}, 2015.

\bibitem{mahajan2011joint_attrib}
D.~Mahajan, S.~Sellamanickam, and V.~Nair, ``A joint learning framework for
  attribute models and object descriptions,'' in \emph{IEEE International
  Conference on Computer Vision}, 2011, pp. 1227--1234.

\bibitem{semantic_graph}
Z.~Fu, T.~Xiang, E.~Kodirov, and S.~Gong, ``zero-shot object recognition by
  semantic manifold distance,'' in \emph{CVPR}, 2015.

\bibitem{Deng2014}
J.~Deng, N.~Ding, Y.~Jia, A.~Frome, K.~Murphy, S.~Bengio, Y.~Li, H.~Neven, and
  H.~Adam, ``Large-scale object classification using label relation graphs,''
  in \emph{ECCV}, 2014.

\bibitem{CCAoverview}
D.~R. Hardoon, S.~Szedmak, and J.~Shawe-Taylor, ``Canonical correlation
  analysis; an overview with application to learning methods,'' in \emph{Neural
  Computation}, 2004.

\bibitem{SocherFeiFeiCVPR2010}
R.~Socher and L.~Fei-Fei, ``Connecting modalities: Semi-supervised segmentation
  and annotation of images using unaligned text corpora,'' in \emph{IEEE
  Conference on Computer Vision and Pattern Recognition}, 2010.

\bibitem{multiviewCCAIJCV}
Y.~Gong, Q.~Ke, M.~Isard, and S.~Lazebnik, ``A multi-view embedding space for
  modeling internet images, tags, and their semantics,'' \emph{International
  Journal of Computer Vision}, 2013.

\bibitem{HwangIJCV}
S.~J. Hwang and K.~Grauman, ``Learning the relative importance of objects from
  tagged images for retrieval and cross-modal search,'' \emph{International
  Journal of Computer Vision}, 2011.

\bibitem{topicimgannot}
Y.~Wang and S.~Gong, ``Translating topics to words for image annotation,'' in
  \emph{ACM International Conference on Conference on Information and Knowledge
  Management}, 2007.

\bibitem{jointly_zsl}
G.-J. Qi, W.~Liu, C.~Aggarwal, and T.~Huang, ``Joint intermodal and intramodal
  label transfers for extremely rare or unseen classes,'' \emph{IEEE TPAMI},
  2017.

\bibitem{szegedy2015going}
C.~Szegedy, W.~Liu, Y.~Jia, P.~Sermanet, S.~Reed, D.~Anguelov, D.~Erhan,
  V.~Vanhoucke, and A.~Rabinovich, ``Going deeper with convolutions,'' in
  \emph{CVPR}, 2015.

\bibitem{deep_0shot_recent}
L.~Zhang, T.~Xiang, and S.~Gong, ``Learning a deep embedding model for
  zero-shot learning,'' in \emph{CVPR}, 2017.

\bibitem{transductiveEmbeddingJournal}
Y.~Fu, T.~M. Hospedales, T.~Xiang, and S.~Gong, ``Transductive multi-view
  zero-shot learning,'' \emph{IEEE TPAMI}, 2015.

\bibitem{synthesized_0shot}
S.~Changpinyo, W.-L. Chao, B.~Gong, and F.~Sha, ``Synthesized classifiers for
  zero-shot learning,'' in \emph{CVPR}, 2016.

\bibitem{fast_0shot}
Y.~Zhang, B.~Gong, and M.~Shah, ``Fast zero-shot image tagging,'' in
  \emph{CVPR}, 2016.

\bibitem{wang2010reg_tag_corr}
Y.~Wang and G.~Mori, ``A discriminative latent model of image region and object
  tag correspondence,'' in \emph{Neural Information Processing Systems}, 2010.

\bibitem{shaoling_cvpr2017}
Y.~Long, L.~Liu, L.~Shao, F.~Shen, G.~Ding, and J.~Han, ``From zero-shot
  learning to conventional supervised classification: Unseen visual data
  synthesis,'' in \emph{CVPR}, 2017.

\bibitem{Eylor_iccv2015}
E.~Kodirov, T.~Xiang, Z.~Fu, and S.~Gong, ``Unsupervised domain adaptation for
  zero-shot learning,'' in \emph{ICCV}, 2015.

\bibitem{transferlearningNIPS}
M.~Rohrbach, S.~Ebert, and B.~Schiele, ``Transfer learning in a transductive
  setting,'' in \emph{NIPS}, 2013.

\bibitem{Li_CVPR2017}
Y.~Li, D.~Wang, H.~Hu, Y.~Lin, and Y.~Zhuang, ``Zero-shot recognition using
  dual visual-semantic mapping paths,'' in \emph{CVPR}, 2017.

\bibitem{zsl_action_xu}
X.~Xu, T.~Hospedales, and S.~Gong, ``Transductive zero-shot action recognition
  by word-vector embedding,'' in \emph{IJCV}, 2016.

\bibitem{marcobaronihubness}
B.~Marco, L.~Angeliki, and D.~Georgiana, ``Hubness and pollution: Delving into
  cross-space mapping for zero-shot learning,'' in \emph{ACL}, 2015.

\bibitem{Low2013}
T.~Low, C.~Borgelt, S.~Stober, and A.~N{\"u}rnberger, \emph{The Hubness
  Phenomenon: Fact or Artifact?}, 2013.

\bibitem{dinu2014improving}
G.~Dinu, A.~Lazaridou, and M.~Baroni, ``Improving zero-shot learning by
  mitigating the hubness problem,'' in \emph{ICLR workshop}, 2014.

\bibitem{shigeto2015ridge}
Y.~Shigeto, I.~Suzuki, K.~Hara, M.~Shimbo, and Y.~Matsumoto, ``Ridge
  regression, hubness, and zero-shot learning,'' in \emph{ECML/PKDD}, 2015.

\bibitem{wild_0shot}
W.-L. Chao, S.~Changpinyo, B.~Gong, and F.~Sha., ``An empirical study and
  analysis of generalized zero-shot learning for object recognition in the
  wild,'' in \emph{ECCV}, 2016.

\bibitem{Scheirer_2014_TPAMIb}
W.~J. Scheirer, L.~P. Jain, and T.~E. Boult, ``Probability models for open set
  recognition,'' \emph{IEEE TPAMI}, 2014.

\bibitem{Scheirer_2013_TPAMI}
W.~J. Scheirer, A.~Rocha, A.~Sapkota, and T.~E. Boult, ``Towards open set
  recognition,'' \emph{IEEE TPAMI}, 2013.

\bibitem{ssvoc_2016_CVPR}
Y.~Fu and L.~Sigal, ``Semi-supervised vocabulary-informed learning,'' in
  \emph{CVPR}, 2016.

\bibitem{ssvoc_evl}
Y.~Fu, H.~Dong, Y.~feng Ma, Z.~Zhang, and X.~Xue, ``Vocabulary-informed extreme
  value learning,'' in \emph{arxiv}, 2017.

\bibitem{zsl_ugly}
Y.~Xian, B.~Schiele, and Z.~Akata, ``Zero-shot learning - the good, the bad and
  the ugly,'' in \emph{CVPR}, 2017.

\bibitem{Bendale_2015_CVPR}
A.~Bendale and T.~Boult, ``Towards open world recognition,'' in \emph{CVPR},
  2015.

\bibitem{Sattar_2015_CVPR}
H.~Sattar, S.~Muller, M.~Fritz, and A.~Bulling, ``Prediction of search targets
  from fixations in open-world settings,'' in \emph{CVPR}, 2015.

\bibitem{gomes2008inc_dpmm}
R.~Gomes, M.~Welling, and P.~Perona, ``Incremental learning of nonparametric
  bayesian mixture models,'' in \emph{IEEE Conference on Computer Vision and
  Pattern Recognition}, 2008.

\bibitem{diehl2003inc_svm}
C.~P. Diehl and G.~Cauwenberghs, ``Svm incremental learning, adaptation and
  optimization,'' in \emph{IJCNN}, vol.~4, 20--24 July 2003, pp. 2685--2690.

\bibitem{iCaRL}
S.~A. Rebuffi, A.~Kolesnikov, G.~Sperl, and C.~H. Lampert, ``icarl: Incremental
  classifier and representation learning,'' in \emph{CVPR}, 2017.

\bibitem{Guadarrama14:OOR}
S.~Guadarrama, E.~Rodner, K.~Saenko, N.~Zhang, R.~Farrell, J.~Donahue, and
  T.~Darrell, ``Open-vocabulary object retrieval,'' in \emph{Robotics Science
  and Systems (RSS)}, 2014.

\bibitem{open_vocab_description}
S.~Guadarrama, E.~Rodner, K.~Saenko, and T.~Darrell, ``Understanding object
  descriptions in robotics by open-vocabulary object retrieval and detection,''
  in \emph{Journal International Journal of Robotics Research}, 2016.

\bibitem{open_world_1shot}
W.~Zheng, S.~Gong, and T.~Xiang, ``Towards open-world person re-identification
  by one-shot group-based verification,'' in \emph{IEEE TPAMI}, 2016.

\bibitem{open_vocab_scen_parsing}
H.~Zhao, X.~Puig, B.~Zhou, S.~Fidler, and A.~Torralba, ``Open vocabulary scene
  parsing,'' in \emph{CVPR}, 2017.

\bibitem{Jankowski}
Jankowski, Norbert, Duch, Wodzislaw, Grabczewski, and Krzyszto, ``Meta-learning
  in computational intelligence,'' in \emph{Springer Science \& Business
  Media}, 2011.

\bibitem{compositional_1shot}
B.~M. Lake and R.~Salakhutdinov, ``One-shot learning by inverting a
  compositional causal process,'' in \emph{NIPS}, 2013.

\bibitem{NIPS2004_2566}
\BIBentryALTinterwordspacing
J.~Goldberger, G.~E. Hinton, S.~T. Roweis, and R.~R. Salakhutdinov,
  ``Neighbourhood components analysis,'' in \emph{Advances in Neural
  Information Processing Systems 17}, L.~K. Saul, Y.~Weiss, and L.~Bottou,
  Eds.\hskip 1em plus 0.5em minus 0.4em\relax MIT Press, 2005, pp. 513--520.
  [Online]. Available:
  \url{http://papers.nips.cc/paper/2566-neighbourhood-components-analysis.pdf}
\BIBentrySTDinterwordspacing

\bibitem{inverse_graphics}
T.~D. Kulkarni, W.~F. Whitney, P.~Kohli, and J.~Tenenbaum, ``Deep convolutional
  inverse graphics network,'' in \emph{NIPS}, 2015.

\bibitem{CAD_models}
T.~D. Kulkarni, V.~K. Mansinghka, P.~Kohli, and J.~B. Tenenbaum, ``Inverse
  graphics with probabilistic cad models,'' in \emph{arxiv:1407.1339}, 2014.

\bibitem{human_level_prob}
B.~Lake, R.~Salakhutdinov, and J.~Tenenbaum, ``Human-level concept learning
  through probabilistic program induction,'' in \emph{NIPS}, 2013.

\bibitem{JVilalta2002AIR}
R.~JVilalta and Y.~Drissi, ``A perspective view and survey of meta-learning,''
  \emph{Artificial intelligence review}, 2002.

\bibitem{feifei2003unsup_1s_objcat_learn}
L.~Fei-Fei, R.~Fergus, and P.~Perona, ``A bayesian approach to unsupervised
  one-shot learning of object categories,'' in \emph{IEEE International
  Conference on Computer Vision}, 2003.

\bibitem{feifei2006one_shot}
------, ``One-shot learning of object categories,'' \emph{IEEE TPAMI}, 2006.

\bibitem{tommasi2009transfercat}
T.~Tommasi and B.~Caputo, ``The more you know, the less you learn: from
  knowledge transfer to one-shot learning of object categories,'' in
  \emph{British Machine Vision Conference}, 2009.

\bibitem{bart2005cross_gen}
E.~Bart and S.~Ullman, ``Cross-generalization: learning novel classes from a
  single example by feature replacement,'' in \emph{CVPR}, 2005.

\bibitem{hertz2016icml}
T.~Hertz, A.~Hillel, and D.~Weinshall, ``Learning a kernel function for
  classification with small training samples,'' in \emph{ICML}, 2016.

\bibitem{Fleuret2005nips}
F.~Fleuret and G.~Blanchard, ``Pattern recognition from one example by
  chopping,'' in \emph{NIPS}, 2005.

\bibitem{amit2007icml}
Y.~Amit, Fink, S.~M., and U.~N., ``Uncovering shared structures in multiclass
  classification,'' in \emph{ICML}, 2007.

\bibitem{wolfc2005cvpr}
L.~Wolf and I.~Martin, ``Robust boosting for learning from few examples,'' in
  \emph{CVPR}, 2005.

\bibitem{torralba2005pami}
A.~Torralba, K.~Murphy, and W.~Freeman, ``sharing visual features for
  multiclass and multiview object detection,'' in \emph{IEEE TPAMI}, 2007.

\bibitem{one_shot_TL_contexutal}
A.~Torralba, K.~P. Murphy, and W.~T. Freeman, ``Using the forest to see the
  trees: Exploiting context for visual object detection and localization,''
  \emph{Commun. ACM}, 2010.

\bibitem{Bromley1993ijcai}
J.~Bromley, J.~Bentz, L.~Bottou, I.~Guyon, Y.~LeCun, C.~Moore, E.~Sackinger,
  and R.~Shah, ``Signature verification using a siamese time delay neural
  network,'' in \emph{IJCAI}, 1993.

\bibitem{siamese_1shot}
G.~Koch, R.~Zemel, and R.~Salakhutdinov, ``Siamese neural networks for one-shot
  image recognition,'' in \emph{ICML -- Deep Learning Workshok}, 2015.

\bibitem{Kienzle2006icml}
W.~Kienzle and K.~Chellapilla, ``Personalized handwriting recognition via
  biased reg- ularization,'' in \emph{ICML}, 2006.

\bibitem{quattoni2008sparse_transfer}
A.~Quattoni, M.~Collins, and T.~Darrell, ``Transfer learning for image
  classification with sparse prototype representations,'' in \emph{IEEE
  Conference on Computer Vision and Pattern Recognition}, 2008, pp. 1--8.

\bibitem{fink2005nips}
M.~Fink, ``Object classification from a single example utilizing class
  relevance metrics,'' in \emph{NIPS}, 2005.

\bibitem{wolf2009iccv}
L.~Wolf, T.~Hassner, and Y.~Taigman, ``The one-shot similarity kernel,'' in
  \emph{ICCV}, 2009.

\bibitem{Weston:2011:WSU:2283696.2283856}
J.~Weston, S.~Bengio, and N.~Usunier, ``Wsabie: Scaling up to large vocabulary
  image annotation,'' in \emph{IJCAI}, 2011.

\bibitem{deep_1shot_recent}
Santoro, S.~Bartunov, M.~Botvinick, D.~Wierstra, and T.~Lillicrap, ``One-shot
  learning with memory-augmented neural networks,'' in \emph{arx}, 2016.

\bibitem{feedforward_1shot}
L.~Bertinetto, J.~F. Henriques, J.~Valmadre, P.~Torr, and A.~Vedaldi,
  ``Learning feed-forward one-shot learners,'' in \emph{NIPS}, 2016.

\bibitem{video2vec_1shot}
A.~Habibian, T.~Mensink, and C.~Snoek, ``Video2vec embeddings recognize events
  when examples are scarce,'' in \emph{IEEE TPAMI}, 2014.

\bibitem{matchingnet_1shot}
O.~Vinyals, C.~Blundell, T.~Lillicrap, K.~Kavukcuoglu, and D.~Wierstra,
  ``Matching networks for one shot learning,'' in \emph{NIPS}, 2016.

\bibitem{infield_1shot}
H.~Zhang, K.~Dana, and K.~Nishino, ``Friction from reflectance: Deep
  reflectance codes for predicting physical surface properties from one-shot
  in-field reflectance,,'' in \emph{ECCV}, 2016.

\bibitem{yuxiong2016eccv}
Y.~Wang and M.~Hebert, ``Learning from small sample sets by combining
  unsupervised meta-training with cnns,'' in \emph{NIPS}, 2016.

\bibitem{yuxiong2016nips}
------, ``Learning to learn: model regression networks for easy small sample
  learning,'' in \emph{ECCV}, 2016.

\bibitem{pmlr-v70-finn17a}
\BIBentryALTinterwordspacing
C.~Finn, P.~Abbeel, and S.~Levine, ``Model-agnostic meta-learning for fast
  adaptation of deep networks,'' in \emph{Proceedings of the 34th International
  Conference on Machine Learning}, ser. Proceedings of Machine Learning
  Research, D.~Precup and Y.~W. Teh, Eds., vol.~70.\hskip 1em plus 0.5em minus
  0.4em\relax International Convention Centre, Sydney, Australia: PMLR, 06--11
  Aug 2017, pp. 1126--1135. [Online]. Available:
  \url{http://proceedings.mlr.press/v70/finn17a.html}
\BIBentrySTDinterwordspacing

\bibitem{rusu-progressive-2016}
A.~A. Rusu, N.~C. Rabinowitz, G.~Desjardins, H.~Soyer, J.~Kirkpatrick,
  K.~Kavukcuoglu, R.~Pascanu, and R.~Hadsell, ``Progressive neural networks,''
  \emph{arXiv preprint arXiv:1606.04671}, 2016.

\bibitem{sift}
D.~G. Lowe, ``Distinctive image features from scale-invariant keypoints,''
  \emph{International Journal of Computer Vision}, vol.~60, 2004.

\bibitem{colorSIFT2008CVPR}
K.~E.~A. van~de Sande, T.~Gevers, and C.~G.~M. Snoek, ``Evaluation of color
  descriptors for object and scene recognition,'' in \emph{IEEE Conference on
  Computer Vision and Pattern Recognition}, 2008.

\bibitem{PHOG2007CVIR}
A.~Bosch, A.~Zisserman, and X.~Munoz, ``Representing shape with a spatial
  pyramid kernel,'' in \emph{ACM International Conference on Image and Video
  Retrieval}, 2007.

\bibitem{bay2008surf}
H.~Bay, A.~Ess, T.~Tuytelaars, and L.~V. Gool, ``{SURF}: Speeded up robust
  features,'' \emph{Computer Vision and Image Understanding}, vol. 110, no.~3,
  pp. 346--359, 2008.

\bibitem{selfsimilarity2007CVPR}
E.~Shechtman and M.~Irani, ``Matching local self-similarities across images and
  videos,'' in \emph{IEEE Conference on Computer Vision and Pattern
  Recognition}, 2007.

\bibitem{decaf2014ICML}
J.~Donahue, Y.~Jia, O.~Vinyals, J.~Hoffman, N.~Zhang, E.~Tzeng, and T.~Darrell,
  ``Decaf: A deep convolutional activation feature for generic visual
  recognition,'' in \emph{International Conference on Machine Learning}, 2014.

\bibitem{WahCUB_200_2011}
C.~Wah, S.~Branson, P.~Welinder, P.~Perona, and S.~Belongie, ``{The
  Caltech-UCSD Birds-200-2011 Dataset},'' California Institute of Technology,
  Tech. Rep. CNS-TR-2011-001, 2011.

\bibitem{scene_OSR}
A.~Oliva and A.~Torralba, ``Modeling the shape of the scene: Aholistic
  representation of the spatial envelope,'' \emph{International Journal of
  Computer Vision}, vol.~42, 2001.

\bibitem{SUN_attrib}
G.~Patterson and J.~Hays, ``Sun attribute database: Discovering, annotating,
  and recognizing scene attributes.'' in \emph{IEEE Conference on Computer
  Vision and Pattern Recognition}, 2012.

\bibitem{xiao2010sunscene}
J.~Xiao, J.~Hays, K.~A. Ehinger, A.~Oliva, and A.~Torralba, ``Sun database:
  Large-scale scene recognition from abbey to zoo,'' in \emph{IEEE Conference
  on Computer Vision and Pattern Recognition}, 2010, pp. 3485--3492.

\bibitem{jiang2011consumervideo}
Y.-G. Jiang, G.~Ye, S.-F. Chang, D.~Ellis, and A.~C. Loui, ``Consumer video
  understanding: A benchmark database and an evaluation of human and machine
  performance,'' in \emph{ACM International Conference on Multimedia
  Retrieval}, 2011.

\bibitem{Zha_ontology}
\BIBentryALTinterwordspacing
Z.-J. Zha, T.~Mei, Z.~Wang, and X.-S. Hua, ``Building a comprehensive ontology
  to refine video concept detection,'' in \emph{Proceedings of the
  International Workshop on Workshop on Multimedia Information Retrieval}, ser.
  MIR '07.\hskip 1em plus 0.5em minus 0.4em\relax New York, NY, USA: ACM, 2007,
  pp. 227--236. [Online]. Available:
  \url{http://doi.acm.org/10.1145/1290082.1290114}
\BIBentrySTDinterwordspacing

\bibitem{deng2009imagenet}
J.~Deng, W.~Dong, R.~Socher, L.-J. Li, K.~Li, and L.~Fei-Fei, ``Imagenet: A
  large-scale hierarchical image database,'' in \emph{CVPR}, 2009.

\bibitem{oxford_flower}
M.-E. Nilsback and A.~Zisserman, ``Automated flower classification over a large
  number of classes,'' in \emph{Proceedings of the Indian Conference on
  Computer Vision, Graphics and Image Processing}, 2008.

\bibitem{ucf101}
K.~Soomro, A.~R. Zamir, and M.~Shah, ``Ucf101: A dataset of 101 human action
  classes from videos in the wild,'' \emph{CRCV-TR-12-01}, 2012.

\bibitem{THUMOS}
H.~Idrees, A.~R. Zamir, Y.-G. Jiang, A.~Gorban, I.~Laptev, R.~Sukthankar, and
  M.~Shah, ``The thumos challenge on action recognition for videos "in the
  wild",'' \emph{Computer Vision and Image Understanding}, 2017.

\bibitem{fcvid_2017}
Y.-G. Jiang, Z.~Wu, J.~Wang, X.~Xue, and S.-F. Chang, ``Exploiting feature and
  class relationships in video categorization with regularized deep neural
  networks,'' in \emph{IEEE TPAMI}, 2017.

\bibitem{activitynet}
F.~C. H. V. E.~B. Ghanem and J.~C. Niebles, ``Activitynet: A large-scale video
  benchmark for human activity understanding,'' in \emph{CVPR}, 2015.

\bibitem{overfeat}
P.~Sermanet, D.~Eigen, X.~Zhang, M.~Mathieu, R.~Fergus, and Y.~LeCun,
  ``Overfeat: Integrated recognition, localization and detection using
  convolutional networks,'' in \emph{ICLR}, 2014.

\bibitem{returnDevil2014BMVC}
K.~Chatfield, K.~Simonyan, A.~Vedaldi, and A.~Zisserman, ``Return of the devil
  in the details: Delving deep into convolutional nets,'' in \emph{BMVC}, 2014.

\bibitem{he2015deep}
K.~He, X.~Zhang, S.~Ren, and J.~Sun, ``Deep residual learning for image
  recognition,'' \emph{arXiv preprint arXiv:1512.03385}, 2015.

\bibitem{GloVec}
J.~Pennington, R.~Socher, and C.~D. Manning, ``Glove: Global vectors for word
  representation,'' in \emph{EMNLP}, 2014.

\bibitem{latent_0shot}
Y.~Xian, Z.~Akata, G.~Sharma, Q.~Nguyen, M.~Hein, and B.~Schiele, ``Latent
  embeddings for zero-shot classification,'' in \emph{CVPR}, 2016.

\bibitem{lifelong_iid}
A.~Pentina and C.~H. Lampert, ``Lifelong learning with non-i.i.d. tasks,'' in
  \emph{NIPS}, 2015.

\bibitem{curriculum_learning}
A.~Pentina, V.~Sharmanska, and C.~H. Lampert, ``Curriculum learning of multiple
  tasks,'' in \emph{CVPR}, 2015.

\end{thebibliography}

\begin{IEEEbiography}{Yanwei Fu}
received the BSc degree in information and computing sciences from Nanjing University of Technology in 2008; and the MEng degree in the Department of Computer Science \& Technology at Nanjing University in 2011, China. He is now pursuing his PhD in vision group of EECS, Queen Mary University of London. His research interest is attribute learning, topic model, learning to rank, video summarization and image segmentation.
\end{IEEEbiography}

\begin{IEEEbiography}{Tao Xiang}
received the Ph.D. degree in electrical and computer engineering from the National University of Singapore in 2002. He is currently a reader (associate professor) in the School of Electronic Engineering and Computer Science, Queen Mary University of London. His research interests include computer vision, machine learning, and data mining. He has published over 140 papers in international journals and conferences.
\end{IEEEbiography}

\begin{IEEEbiography}{Leonid Sigal}
 is an Associate Professor at the University of British Columbia. Prior to this he was a Senior Research Scientist at Disney Research. He completed his Ph.D. at Brown University in 2008; received his M.A. from Boston University in 1999, and M.Sc. from Brown University in 2003. Leonid's research interests lie in the areas of computer vision, machine learning, and computer graphics. Leonid's research emphasis is on machine learning and statistical approaches for visual recognition, understanding and analytics. He has published more than 70 papers in venues and journals in these fields (including TPAMI, IJCV, CVPR, ICCV and NIPS).
\end{IEEEbiography}

\begin{IEEEbiography}{Yu-Gang Jiang}
a Professor in School of Computer Science, Fudan University, China. His Lab for Big Video Data Analytics conducts research on all aspects of extracting high-level information from big video data, such as video event recognition, object/scene recognition and large-scale visual search. His work has led to many awards, including the inaugural ACM China Rising Star Award and the 2015 ACM SIGMM Rising Star Award.
\end{IEEEbiography}

\begin{IEEEbiography}{Xiangyang Xue}
Xiangyang Xue received the B.S., M.S., and Ph.D. degrees in communication engineering from Xidian University, Xi'an, China, in 1989, 1992 and 1995, respectively. He is currently a Professor of Computer Science at Fudan University, Shanghai, China. His research interests include multimedia information processing and machine learning.
\end{IEEEbiography}

\begin{IEEEbiography}{Shaogang Gong}
received the DPhil degree in 1989 from Keble College, Oxford University. He has been Professor of Visual Com- putation at Queen Mary University of Lon- don since 2001, a fellow of the Institution of Electrical Engineers and a fellow of the British Computer Society. His research inter- ests include computer vision, machine learn- ing, and video analysis.
\end{IEEEbiography}

\end{document}